\documentclass{article}

\usepackage[preprint]{neurips_2025}

\usepackage[labelfont=bf]{caption}
\usepackage[utf8]{inputenc} 
\usepackage[T1]{fontenc}    
\usepackage{hyperref}       
\usepackage{url}            
\usepackage{booktabs}       
\usepackage{amsfonts}       
\usepackage{nicefrac}       
\usepackage{microtype}      
\usepackage{xcolor}         
\usepackage{wrapfig}
\usepackage{tabularx}

\usepackage{amsmath}
\usepackage{bm}
\usepackage{graphicx}
\usepackage{caption}
\usepackage{subcaption} 

\usepackage{algorithm2e}
\RestyleAlgo{ruled}
\SetKwComment{Comment}{/* }{ */}

\usepackage{amsmath,amsfonts,bm}

\def\Figref#1{Figure~\ref{#1}}

\def\Secref#1{Section~\ref{#1}}

\def\eqref#1{equation~\ref{#1}}

\def\Algref#1{Algorithm~\ref{#1}}

\def\Appref#1{Appendix~\ref{#1}}

\def\1{\bm{1}}

\def\vpsi{{\bm{\psi}}}
\def\valpha{{\bm{\alpha}}}

\def\vx{{\bm{x}}}
\def\vy{{\bm{y}}}

\def\mA{{\bm{A}}}
\def\mB{{\bm{B}}}

\DeclareMathAlphabet{\mathsfit}{\encodingdefault}{\sfdefault}{m}{sl}
\SetMathAlphabet{\mathsfit}{bold}{\encodingdefault}{\sfdefault}{bx}{n}

\def\sR{{\mathbb{R}}}

\DeclareMathOperator*{\argmin}{arg\,min}

\usepackage[printonlyused]{acronym}
\acrodef{RL}{Reinforcement Learning}
\acrodef{MLP}{Multi-Layer Perceptron}
\acrodef{PPO}{Proximal Policy Optimization}
\acrodef{GAE}{Generalised Advantage Estimation}
\acrodef{MDP}{Markov Decision Process}
\acrodef{AMO}{Average Minimum of the Objective}
\acrodef{ANC}{Average Number of simulator Calls}
\acrodef{ReLU}{Rectified Linear Unit}

\title{Simulating, Fast and Slow: Learning Policies for Black-Box Optimization}

\author{%
  Tim Bakker\thanks{Equal contribution} \\
  Qualcomm AI Research\thanks{Qualcomm AI Research is an initiative of Qualcomm Technologies, Inc.} \\
  \texttt{bakker@qti.qualcomm.com}
  \And
  Fabio Valerio Massoli\footnotemark[1] \\
  Qualcomm AI Research\footnotemark[2] \\
  \AND
  Thomas Hehn \\
  Qualcomm AI Research\footnotemark[2] \\
  \And
  Tribhuvanesh Orekondy \\
  Qualcomm AI Research\footnotemark[2] \\
  \And
  Arash Behboodi \\
  Qualcomm AI Research\footnotemark[2] \\
}

\begin{document}

\maketitle

\begin{abstract}
Simulators are vital in science and engineering, as they faithfully model the influence of design parameters on real-world observations. Optimizing design parameters to minimize a desired objective function is notoriously challenging since simulators are often non-differentiable black boxes. Moreover, each simulation incurs significant compute time, thus doing such optimizations naively may be prohibitively expensive. Furthermore, in experiment design settings, practitioners are typically required to solve the same optimization problem for various input distributions before proceeding to real-world experiments. Starting these related optimizations from scratch each time can be inefficient if the forward simulation model is expensive to evaluate. To address these challenges, this paper introduces a novel method for solving black-box optimization problems on various input distributions by learning a policy that guides the training of a differentiable surrogate. The surrogate's gradients are then used to optimize the simulation parameters with SGD. After training the policy, the cost for downstream optimization of problems involving black-box simulators is amortized, and we require up to $10\times$ fewer expensive simulator calls compared to baselines such as previous local surrogate-based approaches, numerical optimization, and Bayesian optimization.
\end{abstract}

\section{Introduction}

Simulation-based techniques model real-world phenomena (e.g., particle movement, electromagnetic wave propagation) and enable understanding of the influence of system design parameters on resulting observations.
As such, they provide a cheaper alternative to real-world evaluation of system parameters and are invaluable to many fields in physical sciences and engineering, covering domains such as robotics \citep{todorov2012mujoco}, telecommunication \citep{hoydis2023sionna} and particle physics \citep{jonas2019molinv,stakia2021particle}.
Generally, a simulator $f_{\mathrm{sim}}$ models the \textit{forward}-process $f_{\mathrm{sim}}: (\bm{\psi}, \bm{x}) \rightarrow \bm{y}$, which maps simulation parameters $\bm{\psi}$ and input data $\bm{x}$ to observations $\bm{y}$ \citep{shirobokov2020lgso}. 
For example, in particle physics, simulators such as GEANT4 \citep{agostinelli2003geant} and FairRoot \citep{alturany2012} predict the detection of particles $\bm{y}$ given their properties $\bm{x}$, and the properties of the detector $\bm{\psi}$ (e.g., geometry, magnetic field direction and intensity).
Similarly, in wireless communication, simulators such as Matlab RT \citep{matlab2022} or Sionna \citep{hoydis2023sionna} predict the signal strength $\bm{y}$ given scene information $\bm{\psi}$ (e.g., a CAD model of the scene, antenna locations and orientations).

Although simulator design primarily focuses on highly accurate \textit{forward}-models, numerous practical applications require \textit{inverse} inferences. Specifically, inferring unknown system design parameters $\bm{\psi}^*$ that optimize a given objective.
Continuing the previous examples, in the particle physics scenario, $\bm{\psi}$ may characterize the geometrical design of a particle detector to minimize the number of undesired events (background noise).
Similarly, in wireless communications, $\bm{\psi}^*$ may describe the optimal placement of a transmit antenna in a scene to maximize the signal strength across all areas.
Tackling this class of problems using simulators-in-the-loop can be cast as a \textit{black-box optimization problem}: iteratively refining an initial design parameter choice to meet the objective given certain conditions and constraints.
Black-box optimization has rich history, and solutions include gradient-free optimization \citep{banzhaf1998genetic, mahesw2019evo}, Bayesian optimization \citep{oh2018bock, daxberger2020mvbo}, numerical differentiation \citep{alarie2021bbo, shi2023numdiff} or stochastic gradient estimation methods \citep{williams1992reinforce, grathwohl2018void}.
Unlike in typical black-box settings, a major challenge here is that each simulation (for a fixed choice of $\bm{\psi}$) involves significant computation, thus posing a critical bottleneck for iterative optimization. For instance, optimization of particle detectors for detecting rare events in high-energy physics (HEP) experiments is a critical application that demands intensive CPU compute to run simulations. A typical requirement is to study the effect of various radioactive isotopes that could lead to spurious events in the sensitive region of the detector. 
Typically, up to $\sim10^9$ (\citet{massoli2015xenon1t,lantwin2017}) events may be required \emph{per input distribution}. Similarly, in wireless settings, a practitioner may need to run the same simulation multiple times in order to, e.g., find the optimal placement for a transmitting antenna across a variety of frequencies, i.e., varying the input distribution. 
Therefore, finding techniques to amortize these simulation costs is crucial.

\begin{figure}[t]
 \centering  \includegraphics[width=\linewidth]{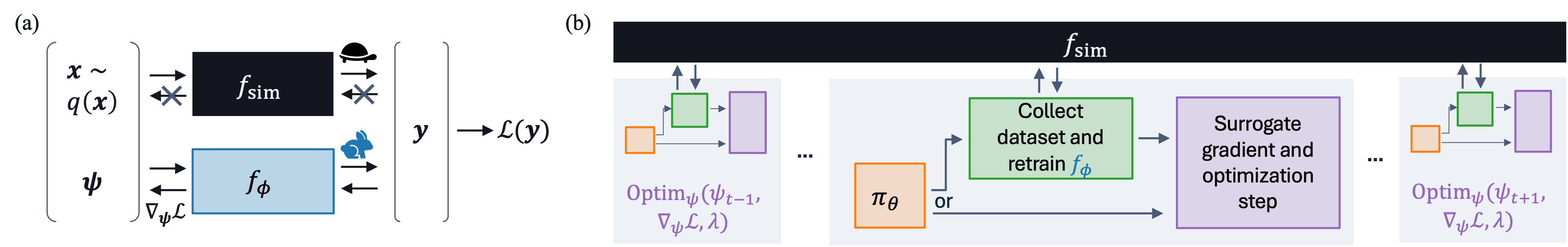}
  \caption{\textbf{Schematic view of our approach}. (a) We study black-box optimization problem (over parameters $\bm{\psi}$), with an emphasis on using gradient information from a fast differentiable surrogate $f_\phi$ (b) To optimize $\bm{\psi}$ sample-efficiently, we employ a policy $\pi_\theta$ to actively determine whether retraining the surrogate is necessary before using the gradient information.}
  \label{fig:approach}
\end{figure}

In this paper, we focus on stochastic gradient estimation techniques for black-box optimization. Inspired by \citet{shirobokov2020lgso}, our approach involves leveraging gradients from a surrogate model trained to (locally) mimic the black-box simulator\footnote{The simulator can be either stochastic or deterministic.}. 
Gradient-based methods typically perform many simulator calls to estimate gradients and are thus computationally demanding. To mitigate these demands, we propose to minimize the number of required simulator calls by learning a \emph{policy} to guide the optimization. The policy determines whether the current surrogate model (fast, but potentially inaccurate) can be used for gradient estimation or whether a simulator call (slow, but more accurate) is necessary to update the surrogate instead, see~\Figref{fig:approach}. Drawing inspiration from the literature on active learning \citep{lal2015orig, lal2017, lal2017dqn, ravi2018meta, lal2018metarl, lal2018imit, lal2018uci, bakker2023lal}, we also let our policy learn how to sample new data for training the local surrogate model, offering additional control. 
Although policy training requires an initial computational investment, a trained policy reduces the number of required simulator calls for downstream optimization problems; that is, training a policy amortizes downstream costs for settings where practitioners need to solve similar optimization problems
many times (e.g., physics and wireless problems). 

Our contribution can be summarized as follows: 
(i) We introduce an \ac{RL} framework for training a policy to reduce the number of computationally expensive calls to a black-box simulator required to solve optimization problems; 
(ii) We propose to learn a policy that determines \emph{when} a simulator call is necessary to update the surrogate and when the current surrogate model can be used instead; 
(iii) We implement a policy that, in addition, learns \emph{how} to sample new data for training the surrogate model during the optimization process; 
(iv) We assess the benefits of our RL-based approach on low- and high-dimensional global optimization benchmark functions and two real-world black-box simulators. Once trained, our policy reduces the number of simulator calls required for optimization on related input distributions up to $10\times$, compared to the baselines.
\section{Related work}\label{sec:rel_works}

\textbf{Black-box optimization} problems are ubiquitous in science and engineering, encompassing scenarios where unknown parameters must be deduced from observational data. These parameters can entail anatomies in MRI \citep{zbontar2018fastmri}, molecular structures \citep{jonas2019molinv}, particle properties \citep{agostinelli2003geant, stakia2021particle}, and cosmological model parameters \citep{cole2022sbi}, among others. The forward process is often a complex physical process that can be modeled by a simulator but does not provide a likelihood for easy inference. Simulation-based inference techniques aim to infer posterior distributions over these simulation parameters in such likelihood-free settings \citep{brookes2019sbilite, cranmer2020sbi, cole2022sbi}. Other solutions may involve supervised learning on observation-parameter pairs or imitation learning \citep{jonas2019molinv, sriram2020varnet}.

Our \textbf{simulator-based optimization} setting is a variation on these problems. Here, the objective is to find the optimal parameters of the simulator, where optimality is typically formulated in terms of desired observations. This methodology can be applied in various fields, such as MRI \citep{pineda2020active, bakker2020greedy, bakker2022multi}, particle physics \citep{stakia2021particle, fanelli2022ioncollider, dorigo2023diff, gorordo2023llp} and molecular design \citep{schwalbekoda2021advmol}. When the simulators are differentiable, direct gradient-based optimization can perform well \citep{avila2018diffphys, hu2019robots, degrave2019diffrob}. However, a different approach is necessary in cases where the simulators are non-differentiable. Well-known gradient-free methods that may be employed in such settings include evolutionary strategies \citep{banzhaf1998genetic, hansen2006cma, mahesw2019evo} and Bayesian optimization \citep{frazier2018tutorial, oh2018bock, eriksson2019bo, daxberger2020mvbo}. 

With the rise of deep learning, there has been a surge of interest in \textbf{approximate-gradient optimization} methods. While some authors consider numerical differentiation \citep{alarie2021bbo, shi2023numdiff}, many others have focused on methods for efficiently obtaining approximate stochastic gradients \citep{williams1992reinforce, grathwohl2018void, ruiz2018learning, mohamed2020mcge, louppe2019advaropt, agrawal2023multifidelity}. Another strategy involves training differentiable surrogate models to mimic the simulator and assuming that the gradients of the surrogate model are similar enough to those of the simulator to be useful for use with stochastic gradient algorithms \citep{shirobokov2020lgso}. Alternatively, one may attempt to directly learn a mean gradient using numerical differences \citep{sarafian2020egl}. Surrogate models have been trained for many applications, including wireless propagation modeling \citep{levie2021radiounet, orekondy2023winert}, space weather prediction \citep{baydin2023space}, material discovery \citep{deepmind2023gnome}, and fluid dynamics simulation \citep{agrawal2024probabilistic}. This trend provides an opportunity for surrogate-based optimization of simulators, as surrogate models are readily available. Additionally, it has been observed by \citet{shirobokov2020lgso} that using (local) surrogate gradients is more efficient than many alternatives. Our work generalizes this setup by introducing a policy that guides the optimization by suggesting when and, optionally, how the surrogate should be updated during the optimization process.
\section{Background} \label{sec:background}

We aim to optimize the simulation parameters of a black-box simulator using stochastic gradient descent (SGD). 
The black-box simulator, $f_{\mathrm{sim}}$, describes a stochastic process\footnote{A non-stochastic simulator can be considered as a special case where $f_{\mathrm{sim}}$ places a delta distribution over observations.}, $p(\bm{y}|\bm{\psi}, \bm{x})$, from which we obtain the observations as $\bm{y}\ =\ f_{\mathrm{sim}}(\vpsi, \vx) \sim p(\bm{y}|\bm{\psi}, \bm{x})$, where $\bm{x} \sim q(\bm{x})$ is a stochastic input (e.g., particle energies) and $\vpsi$ is the vector of simulation parameters. 
Since these simulators are typically not differentiable, we train a surrogate neural network to locally (in $\bm{\psi}$) approximate the simulator \citep{shirobokov2020lgso}. Gradients of these local surrogates, obtained through automatic differentiation, may then be used to optimize $\bm{\psi}$. 
The goal is now to minimize an expected loss $\mathcal{L}$ over the space of the simulation parameters $\bm{\psi}$. As the functional form of the simulator is generally unknown, this expectation cannot be evaluated exactly and is instead estimated using $N$ Monte Carlo samples:
\begin{align}
    \bm{\psi}^* &= \argmin_{\bm{\psi}} \mathbb{E} \left[ \mathcal{L}(\bm{y}) \right] \\ 
    &= \argmin_{\bm{\psi}} \int \mathcal{L}(\bm{y}) \, p(\bm{y}|\bm{\psi}, \bm{x}) \, q(\bm{x}) \, d\bm{x} \, d\bm{y} \\
    &\approx \argmin_{\bm{\psi}} \, \frac{1}{N} \, \sum_{i=1}^N \mathcal{L}(f_{\mathrm{sim}}(\bm{\psi}, \bm{x}_i))
\end{align}
After training a neural network surrogate $f_\phi: (\bm{\psi}, \bm{x}, \bm{z}) \rightarrow \bm{y}$ on data generated with $f_{\mathrm{sim}}$, the optimization might be performed following gradients of the surrogate. Here, $\bm{z}$ is a random latent vector that accounts for the stochasticity of the simulator. Gradients are then estimated as follows:
$ 
    \nabla_{\bm{\psi}} \mathbb{E} \left[ \mathcal{L}(\bm{y}) \right] \approx \frac{1}{N} \, \sum_{i=1}^N \, \nabla_{\bm{\psi}} \, \mathcal{L} \left( f_\phi(\bm{\psi}, \bm{x}_i, \bm{z}_i) \right).
$ 
\citet{shirobokov2020lgso} propose to optimize $\bm{\psi}$ through gradient descent, using these gradients estimated by local surrogates at every descent step. However, since running the forward process $f_{\mathrm{sim}}$ is computationally expensive, training the surrogate for every gradient step is expensive. In this work, we aim to reduce the number of simulation steps required for this optimization.

\section{Policy-based Black-Box Optimization} \label{sec:formulation}
Following \citet{shirobokov2020lgso}, we also perform an iterative optimization based on the surrogate gradients obtained in~\autoref{sec:background}. Starting from an initial parameter $\bm{\psi}_0$, we may train a local surrogate for the current parameter value. To that aim, we start by sampling nearby parameter values within a so-called trust-region, a box of fixed size $2\epsilon$, centered around the current $\bm{\psi}$: $U_\epsilon^{\bm{\psi}} = \{\bm{\psi}'; |\bm{\psi }'-\bm{\psi}|\leq \epsilon\}$. These are combined with input samples from $q(\vx)$ and noise $\bm{z}$, and the simulator $f_\phi(\bm{\psi}, \bm{x}, \bm{z})$ is called to obtain the corresponding $\bm{y}$ values. The surrogate is then trained on these local samples. To improve data efficiency and to allow surrogates to continuously improve in difficult regions, the samples are also stored in a history buffer $H$. Then, surrogate training for a point $\bm{\psi}$ reuses all datapoints in $H$ that lie within $U_\epsilon^{\bm{\psi}}$. The overall process required to generate new samples from the black-box simulator is what we refer to as a ``\emph{simulator call}''. Such a simulator call requires running the forward process on $10^4$ to $10^5$ total $(\bm{\psi}, \bm{x})$ tuples, depending on the precise setting.




\begin{figure}[ht]
  \begin{minipage}[c]{0.48\linewidth}
    \includegraphics[width=\linewidth]{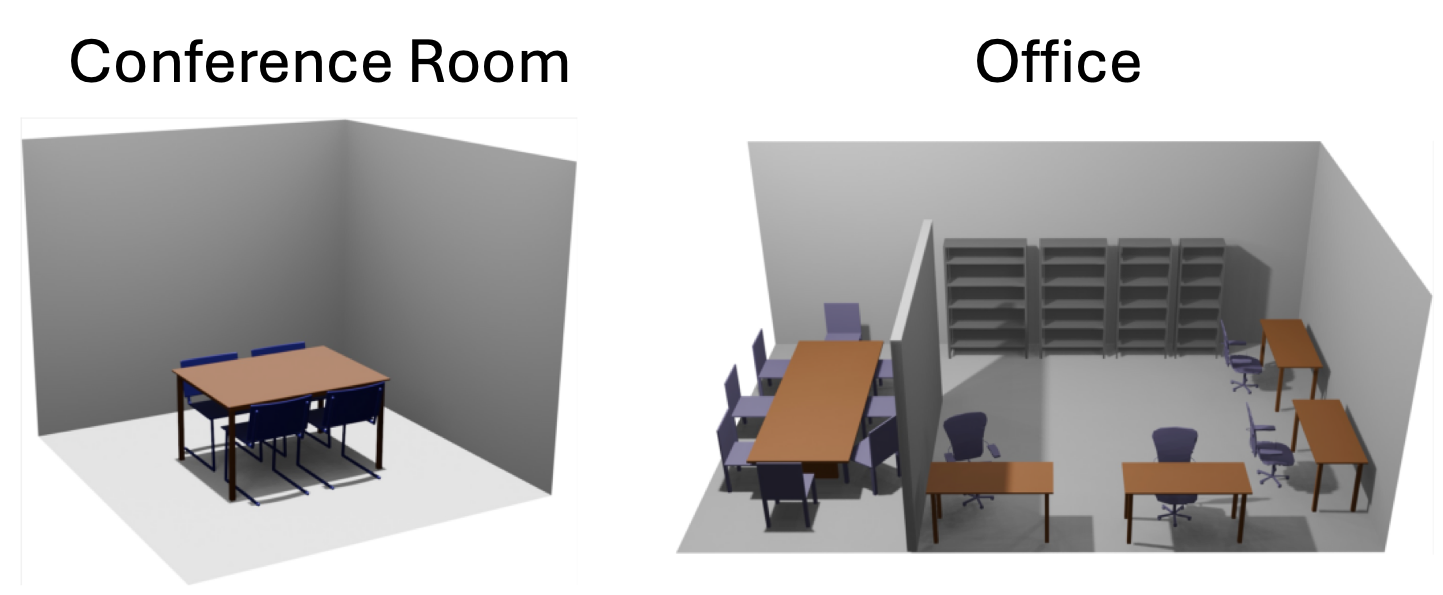}
    \caption{\textbf{3D rendering}. Indoor environments for wireless simulations.}
    \label{fig:wireless_rooms}
  \end{minipage}
  \hfill
  \begin{minipage}[c]{0.48\linewidth}
    \includegraphics[width=\linewidth]{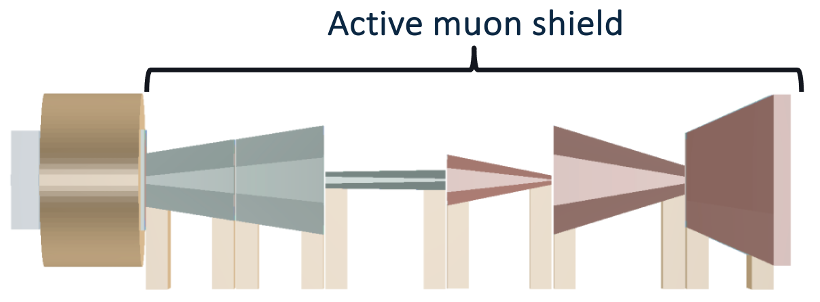}
    \caption{\textbf{Muon shield schematic view}. Baseline configuration of the active muon shield used for physics simulations (image from~\citep{baranov2017optimising}. IOP Publishing, 2017. Licensed under CC BY 3.0).}
    \label{fig:ship_pic}
  \end{minipage}%
\end{figure}

\paragraph{Policy-based Approach} We propose to reduce the number of simulator calls required for an optimization run with an \ac{RL}-based approach. Our method involves utilizing a learned policy $\pi_\theta$, with learnable parameters $\theta$ to: i) decide whether a simulator call should be performed to retrain the local surrogate; and ii) define how to sample surrogate training data from the black-box simulator. This allows surrogates to be retrained -- which involves expensive simulator calls -- only when (and how) the policy chooses it. Thus, if the policy considers the current local surrogate reliable, we can use it for gradient estimation and avoid training a new local surrogate for this purpose.

\paragraph{States} We formalize the sequential optimization process as an episodic \ac{MDP}. The state $s_t$ at timestep $t$ is given by the tuple $(\bm{\psi}_t, t, l_t, \sigma_t)$, where $\bm{\psi}_t$ is the current parameter value, $l_t$ is the number of simulator calls already performed in the episode, and $\sigma_t$ is some measure of uncertainty produced by the surrogate. These features allow the policy to decide based on how long the optimization has been going for and how reliable the surrogate estimates are -- see Appendix~\ref{sup:pomdp} for a discussion regarding observability in the \ac{MDP}.

\paragraph{Actions} Actions $a_t$ consist of binary valued variables $b \in \{0, 1\}$, sampled from a Bernoulli distribution, where $1$ represents the decision to perform a simulator call. We also train policies to output, as part of the action, the trust region size $\epsilon$ for sampling new values for $\bm{\psi}$. State transitions of the \ac{MDP} are by the Adam optimizer~\citep{kingma2014adam}, which updates the current state by doing a single step in the direction of the surrogate gradient estimate. 

\paragraph{Reward design} Episodes end under one of three conditions: $\mathcal{A)}$ the optimization reaches a parameter for which $\mathbb{E} \left[ \mathcal{L}(\bm{y}) \right]$ is below the target value $\tau$ (we call this \textit{termination} - see~\Appref{sup:term_val} for details regarding the choice of $\tau$); $\mathcal{B)}$ the maximum number of timesteps $T$ is reached; or $\mathcal{C)}$ the policy hits the available budget for simulator calls $L$. To incentivize reducing the number of simulator calls, rewards $r(s_t, a_t, s_{t+1})$ are $0$ if $b=0$ and $-1$ if $b=1$. In addition, a reward penalty is added when $\mathcal{B)}$ or $\mathcal{C)}$ occurs to promote termination. The penalty is $-(L-l_t)-1$ when $\mathcal{B)}$ occurs and $-1$ when $\mathcal{C)}$ occurs. This ensures the sum-of-rewards for non-terminating episodes is $-L-1$. We have observed that using reward penalties based on $l_t$ rather than $t$ improves training stability -- see~\Appref{sup:rew_design} for further details concerning the reward design. 

\paragraph{Local Surrogate}
Our local surrogates are implemented as a two-layer \ac{MLP} with \ac{ReLU} activation function. They are trained for three epochs with batch size 512 on $10^4$ to $10^5$ simulator samples, depending on the exact setting (Appendix~\ref{sup:surrogate}). \footnote{Note that the number of simulator forward passes per ``\emph{simulator call}'' is a hyperparameter that we adopted as-is from \citet{shirobokov2020lgso}.} These surrogates are a dramatic simplification from the GAN~\citep{goodfellow2014gan} architecture proposed in \citet{shirobokov2020lgso}. Training GANs is notoriously more challenging than training a shallow \ac{MLP} due to instabilities and mode collapse. Nonetheless, our simpler surrogate can approximate highly complex stochastic and non-stochastic simulators locally. Our policy's decision to retrain a local surrogate -- and thus perform a simulator call -- should rely on the surrogate's reliability. A well-fitted surrogate to the simulator at the current $\bm{\psi}$ will presumably provide useful gradients, so gathering additional data and retraining is unnecessary. Vice-versa, a poorly fitted surrogate will likely not provide useful gradients and may be worth retraining, even if a simulator call is expensive. Determining whether the surrogate is reliable is not straightforward using the available features. To simplify this task, we give the policy an extra feature $\sigma$ that estimates the uncertainty in the surrogate predictions. 

\paragraph{Local Surrogate Ensemble} To construct $\sigma$, we create an ensemble of \emph{three} local surrogates, all trained on and applied to the same input data. We compute the mean prediction per surrogate on $D=100$ input samples as $\bm{\bar{y}} = \frac{1}{D} \, \sum_{i=1}^D \left[ f_\phi(\bm{\psi}, \bm{x}_i, \bm{z}_i) \right]$ for the current $\bm{\psi}$, and construct $\sigma$ as the standard deviation over these three mean predictions. The use of an ensemble empowers our approach with the ability to estimate uncertainties while avoiding the need to train a Bayesian posterior network~\citep{lakshminarayanan2017simple,fort2019deep,wilson2020bayesian}. Since our surrogates are small \ac{MLP}s, the additional resource requirement for training an ensemble instead of a single surrogate is negligible. Gradient steps in $\vpsi$ for simulator optimization are taken using the average gradient estimated from the ensemble -- see~\Appref {sup:surrogate} for further details concerning model implementation.

\paragraph{Policy Training} The policy comprises two separate neural networks: an Actor and a Critic. Both networks are \ac{ReLU} \ac{MLP}s with a single hidden layer of 256 neurons. The input to both networks is the state tuple: $(\bm{\psi}_t, t, l_t, \sigma_t)$. We train our policy in an episodic manner by accumulating sequential optimization episodes and updating the policy using \ac{PPO} \citep{schulman2017ppo} with \ac{GAE} advantages \citep{schulman2016gae} -- see Appendix~\ref{sup:policy} for implementation details and~\ref{sup:training_details} for training details.
\section{Experimental Results} \label{sec:exp_res}

\begin{figure}[ht]
  \begin{subfigure}[h]{0.49\linewidth}
    \includegraphics[width=\linewidth]{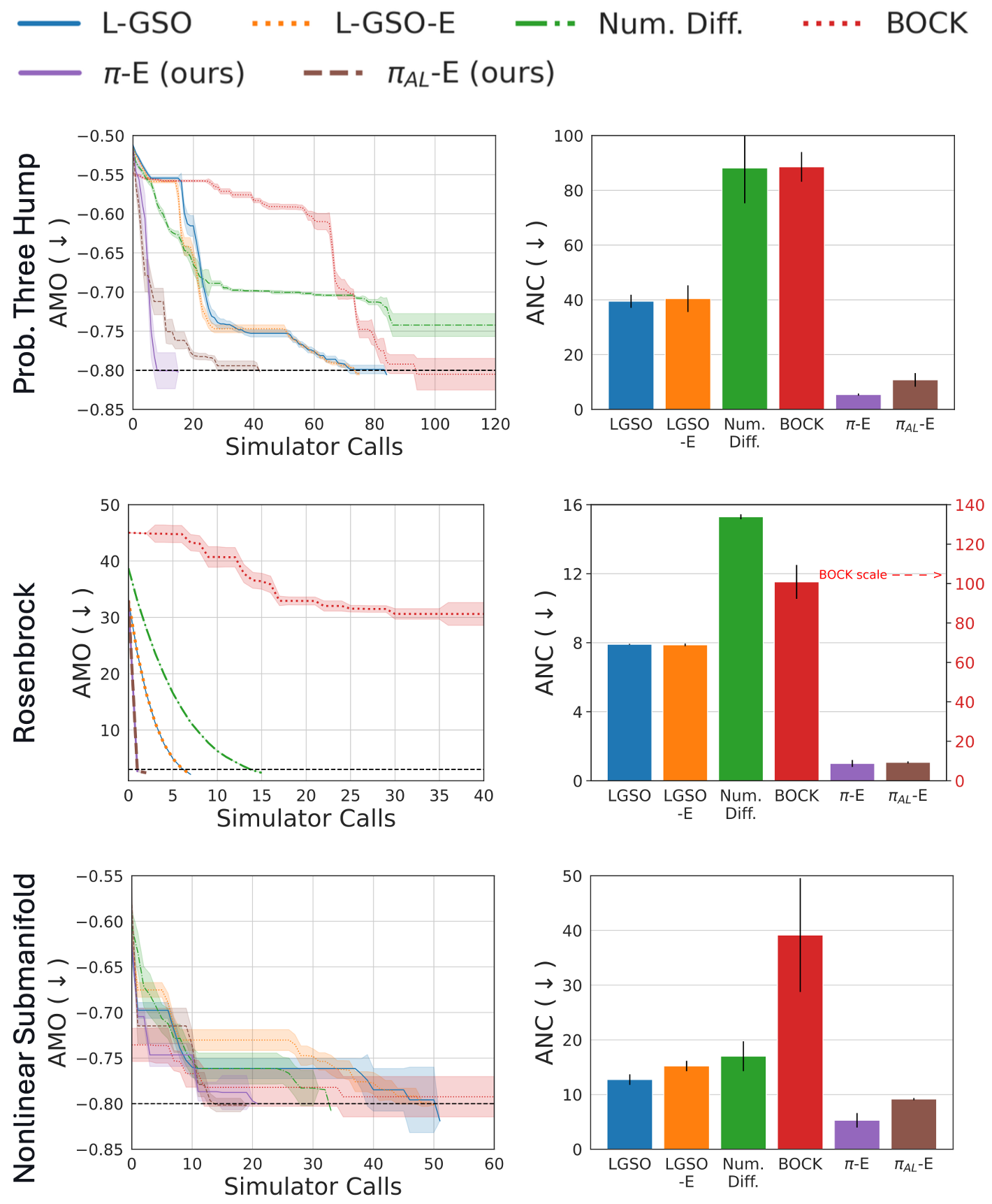}
    \caption{Fixed $\vx$-distribution.}
    \label{fig:benchmark_function_results_fixed_x_dist}
  \end{subfigure}
  \hfill
  \begin{subfigure}[h]{0.5\linewidth}
    \includegraphics[width=\linewidth]{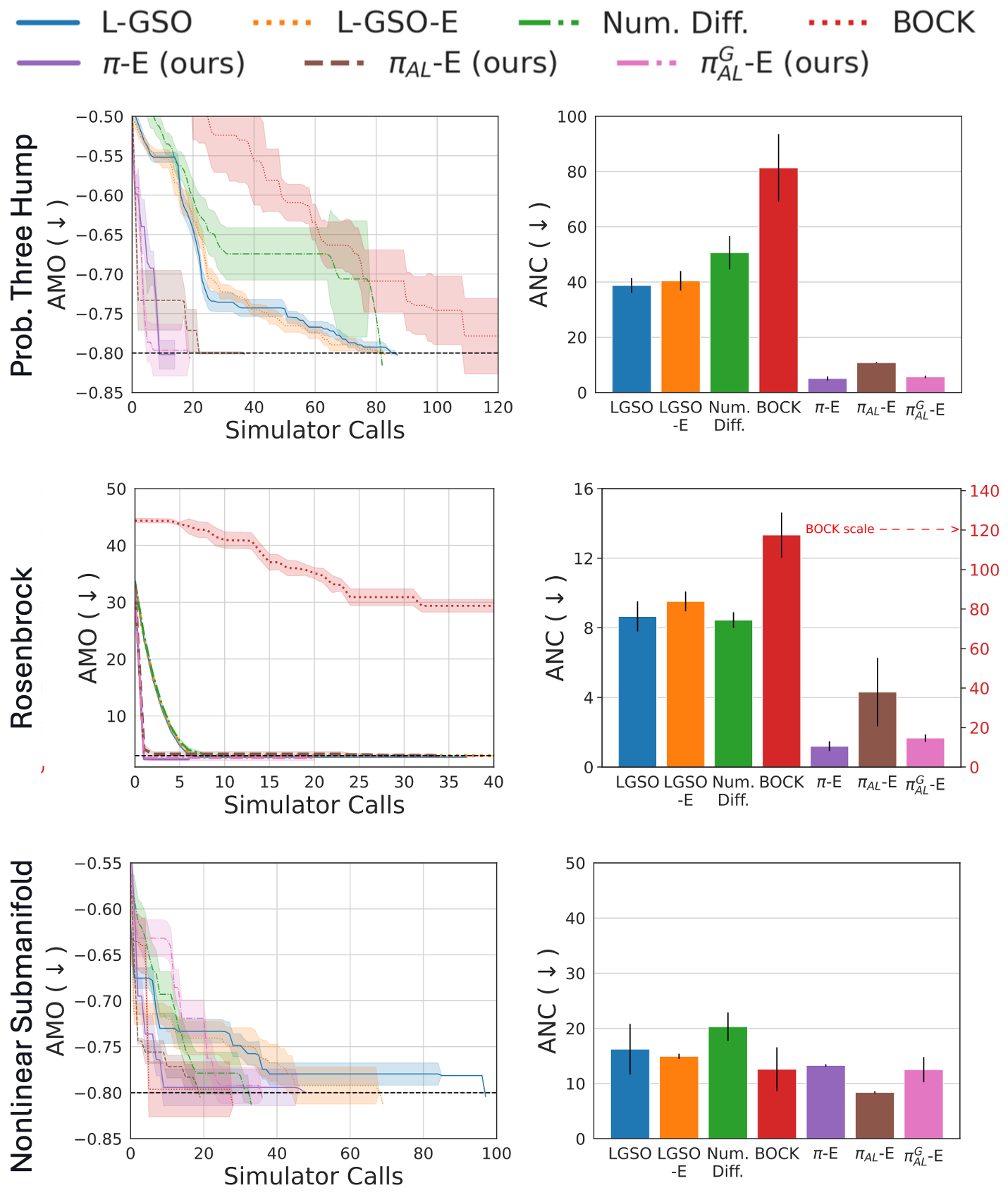}
    \caption{Varying $\vx$-distribution.}
    \label{fig:benchmark_function_results_varying_x_dist}
  \end{subfigure}%
  \caption{Benchmark function results. \textbf{Top} row: Probabilistic Three Hump problem. \textbf{Middle} row: Rosenbrock problem. \textbf{Bottom} row: Nonlinear Submanifold Hump Problem.}
\end{figure}

We evaluate the performance of our method on two sets of simulators. The first set consists of stochastic benchmark functions from the optimization literature~\citep{jamil2013literature}. We consider the Probabilistic Three Hump, Rosenbrock, and Nonlinear Submanifold Hump problems. These benchmark functions are relevant for two reasons:
(i) they allow us to compare our models against baselines on similar settings as in~\citep{shirobokov2020lgso}; and
(ii) they are relatively cheap, allowing us to gain insights into model performance quickly. Since the Probabilistic Three Hump problem is two dimensional, i.e.~$\vpsi \in \mathbb{R}^2$, we are conveniently able to visualize the objective landscape as well as the optimization trajectories. Furthermore, the Rosenbrock and Nonlinear Submanifold Hump problems allow us to test our approach on high-dimensional, more complex problems before moving to real-world black-box simulators. The second type of experiment is on real-world black-box simulators. We consider applications from two different scientific fields: the \textit{Indoor Antenna Placement} problem for wireless communications and the \textit{Muon Background Reduction} problem for high-energy physics.

\paragraph{Baselines}
We compare our method against three baselines. We consider Bayesian optimization using Gaussian processes with cylindrical kernels~\citep{oh2018bock}, numerical differentiation with gradient descent, and local surrogate-based methods (L-GSO)~\citep{shirobokov2020lgso}. Furthermore, to guarantee a fair comparison against our models, we have formulated an ensemble version\footnote{L-GSO-E averages the gradients over the ensemble the same way our method does. The model does not leverage uncertainty since it always calls the black-box simulator.} (L-GSO-E) of the local surrogate for L-GSO.

\paragraph{Our Models}
Policy methods are split into those that only output \textit{when} to perform a simulator call, $\pi$-E, and those that also output \textit{how} to sample $\vpsi$ values (for surrogate training) by providing the neighbour size $\epsilon$ that parameterizes the acquisition function $\pi_{AL}$-E. L-GSO, its ensemble version (L-GSO-E), and $\pi$-E use a fixed value for $\epsilon$ that depends on the problem (see~\Appref{sup:sim}).
$\pi_{AL}^G$-E is a version of $\pi_{AL}$-E where the surrogate ensemble is always warm-started from the previous training step, such that the surrogate is continuously improved along the observed trajectories through $\bm{\psi}$-space (see~\Appref{sup:policy} for more details). Such an approach is potentially more data efficient for surrogate training and may lead to a semi-local surrogate that approximates the simulator along entire optimization paths seen during training. All policy models are trained on-policy using \ac{PPO} on full optimization episodes. In the varying $\vx$-distribution setting, each episode is initialized with a separately sampled $\vx$-distribution, and thus a separate function to be optimized; see Appendix~\ref{sup:param_input} for further details. Metrics shown in this section are computed on trained policies.

\paragraph{Metrics}
We report experimental results on two metrics: the Average Minimum of the Objective function (AMO) for a specific budget of simulator calls and the Average Number of simulator Calls (ANC) required to terminate an episode. The first quantity tells us the objective function value achievable as a function of simulator calls; this is an indicator of the efficacy of each simulator call. The second quantity tells us the average simulator call budget required to solve a black-box optimization problem; this indicates how well the policy leverages the surrogate during optimization. For more details, see \Appref{sup:metrics}. In all experiments, expectations and standard error are quantified over evaluation episodes with three different random seeds.

\subsection{Benchmark Functions}
We experiment with fixed and varying input distributions for each benchmark function. The latter setup solves an entire family of related optimization problems, each characterized by a different input distribution, $q(\vx)$. In particular, during training of the policy and evaluation of all methods for a given simulator, each function to optimize is induced by a different input distribution. We include a short description of our benchmark functions below; more details -- including those on how the input distributions vary -- can be found in~\Appref{sup:sim} and~\Appref{sup:param_input}.

\paragraph{Probabilistic Three Hump Problem}
In the Probabilistic Three Hump problem, the goal is to find the 2-dimensional $\vpsi^*$ such that: $\vpsi^* = \mathrm{arg\ min}_\vpsi\ \mathbb{E}[\sigma(y-10)-\sigma(y)]$, where $\sigma$ is the sigmoid function and $\vy$, the observation vector, is sampled from a distribution that depends on $\vx$ and $\vpsi$.
Since this function is two-dimensional in $\vpsi$, its optimization trajectory is amenable to visualization. \Figref{fig:main_three_hump_psi_opt_traj} of Appendix~\ref{sup:ppo} illustrates such a trajectory, where a fully trained policy heavily exploits the local surrogate for gradient estimation, only performing simulator calls when the model is far from the initial training location (the red square in~\Figref{fig:main_three_hump_psi_opt_traj} denotes the initial trust region $U_{0.5}^{\vpsi_0}$). Intuitively, we expect the surrogate model to provide meaningful gradients in proximity to the $\vpsi$ region where it was previously trained. As the optimization moves away from that region, we expect the quality of the gradients to decline until a simulator call is triggered and the local surrogate is re-trained. Of course, moving away from the last training region is not the only condition that may trigger a simulator call. For instance, towards the end of the trajectory, the policy calls the simulator twice to gather more data and update the surrogate. It may, for instance, have learned that a rapidly changing loss landscape should trigger a simulator call; these nuances are up to the policy to learn.

\paragraph{Rosenbrock Problem}
The aim is to optimize $\vpsi \in \mathbb{R}^{10}$ such that: $\vpsi^* =\mathrm{arg\ min}_\vpsi\ \mathbb{E}[y]$; where $y$ is given by: $y \sim \mathcal{N} (y;\ \gamma+ x,\ 1)$, with $\gamma = \sum^{n-1}_{i=1} \left[ (\psi_i-\psi_{i+1})^2 + (1 - \psi_i )^2 \right]$. This is a higher-dimensional benchmark function that differs in character from the Three Hump problem, allowing us to check that our design is not overoptimized to the Three Hump case.


\paragraph{Nonlinear Submanifold Hump Problem}
This problem is formulated similarly to the Probabilistic Three Hump problem. We consider the embedding $\hat{\vpsi}=\mB\ \mathrm{tanh}(\mA \vpsi)$, where $\mA \in \mathbb{R}^{16 \times 40}$ and $\mB \in \mathbb{R}^{2 \times 16}$, of the vector $\vpsi \in \mathbb{R}^{40}$. Subsequently, $\hat{\vpsi}$ is used in place of $\vpsi$ in the Probabilistic Three Hump problem definition. This essentially embeds the 2-dimensional surface into a 40-dimensional space, allowing us to test the policy's performance on higher-dimensional problems.


\subsection{Real-world Simulators}\label{sec:real_w_s}
We now focus on real-world optimization problems involving computationally expensive, non-differentiable black-box simulators. First, from the field of wireless communications, we consider two settings with a (non-stochastic) wireless ray tracer~\citep{matlab2022}. Second, we move on to the world of subatomic particles and solve a detector optimization problem for which we use Geant4~\citep{agostinelli2003geant} (stochastic simulator) and FairRoot~\citep{alturany2012}. 


\paragraph{Wireless Communication: Indoor Transmitting Antenna Placement}
We study the problem of optimally placing a transmitting antenna in indoor environments to maximize the signal strength at multiple receiver locations.
Determining the signal strength in such a scenario requires a wireless ray tracer (\citet{matlab2022} in our case), which takes as input the transmit location candidate $\vpsi \in \sR^3$ alongside other parameters (e.g., receiver locations, 3D mesh of scene).
To predict the signal strength for a particular \textit{link} (i.e., a transmit-receive antenna pair), the ray tracer exhaustively identifies multiple propagation paths between the two antennas and calculates various attributes of each path (e.g., complex gains, time-of-flight).
The signal strength is computed from the coherent sum of the complex-valued gains of each path impinging on the receive antenna and is represented in log-scale (specifically, dBm).
Optimally placing the transmit antenna is typically slow, as it amounts to naively and slowly sweeping over transmit location choices $\vpsi$ and observing the simulated signal strengths.
Instead, we employ our approach to ``backpropagate'' through the surrogate and perform gradient descent steps on the location $\vpsi$.
Specifically, we consider two indoor scenes for this experiment and investigate how to use our approach to find an optimal transmit location that maximizes signal strength in the 3D scene (\Figref{fig:wireless_rooms}).
The goal in both cases is to find an optimal transmit antenna location $\vpsi$ that maximizes the median signal strength calculated over a distribution of receive locations $\vx \sim q(\vx)$ (see~\Appref{sup:real_world_sim} for more details regarding simulations).
\paragraph{Physics: Muon Background Reduction}
We consider the optimization of the active muon shield for the SHiP experiment~\citep{lantwin2017}. 
Optimizing a detector is a crucial step in designing experiments for particle physics. Detector geometry, intensity and orientation of magnetic fields, and materials used to build the detector all play a crucial role in defining the detector's ``sensitivity'' to specific types of particle interactions, i.e. events, and may need to be optimized.
Observed events are usually divided into signal -- interactions physicists are interested in studying -- and ``background'' -- events that are not of any interest and that might reduce the detector's sensitivity. In the SHiP experiment, muons represent a significant source of background, and it is thus necessary to shield the detector against those particles.
The shield comprises six magnets (\Figref{fig:ship_pic}) each described by seven parameters; hence, $\vpsi \in \mathbb{R}^{42}$. We use the Geant4~\citep{agostinelli2003geant} and FairRoot~\citep{alturany2012} toolkits to run the simulations. The input distribution $\vx$ describes the properties of incoming muons\footnote{For the muon distribution, we use the same dataset as~\citet{shirobokov2020lgso}, which is available for research purposes.}. Specifically, as in~\citep{shirobokov2020lgso}, we consider the momentum ($P$), the azimuthal ($\phi$) and polar ($\theta$) angles with respect to the incoming $z$-axis, the charge $Q$, and $(x, y, z)$ coordinates. The goal is to minimize the expected value of the following objective function: 
$ 
\mathcal{L}(\vy;\valpha) = \sum^{N}_{i=1} \mathbb{I}_{Q_i=1}\sqrt{(\alpha_1 - (\vy_i + \alpha_2))/\alpha_1} + \mathbb{I}_{Q_i=-1}\sqrt{(\alpha_1 + (\vy_i - \alpha_2))/\alpha_1},
$ 
where $\mathbb{I}$ is the indicator function, $\alpha_1$ and $\alpha_2$ are known parameters defining the sensitive region of the detector, and $Q$ and $\vy$ represent the electric charge and the coordinates of the observed muons, respectively. This corresponds to minimizing the number of muons hitting the sensitive region of the detector.

\subsection{Results \& Discussion}



\begin{figure}[ht]
  \begin{minipage}[b]{0.49\linewidth}
    \includegraphics[width=\linewidth]{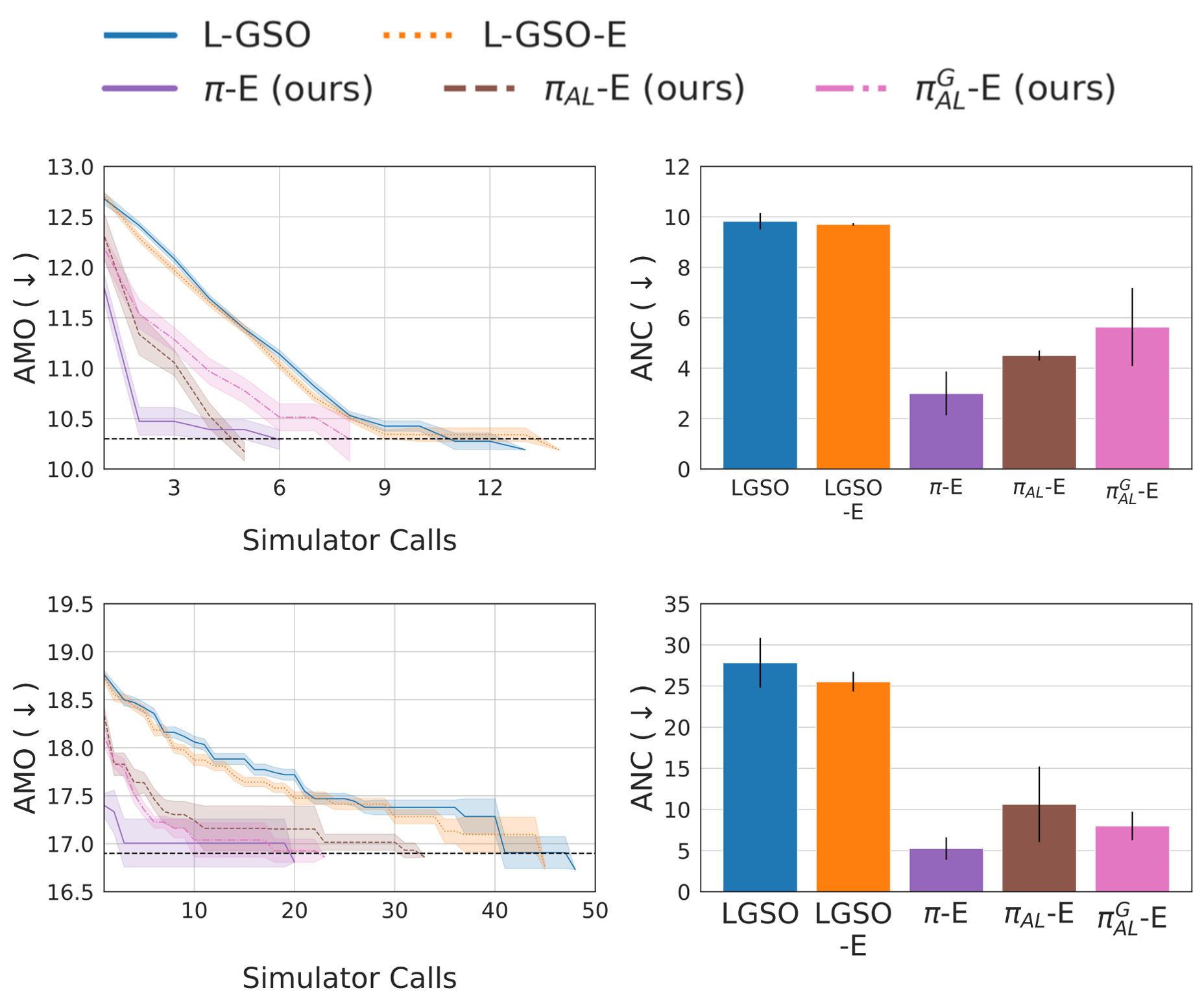}
    \caption{\textbf{Wireless ray-tracing} results for fixed $\vx$-distribution. \textbf{Top} row: Conference room environment. \textbf{Bottom} row: Office room environment.}
    \label{fig:wireless_fixed_x_dist}
  \end{minipage}
  \hfill
  \begin{minipage}[b]{0.49\linewidth}
    \raisebox{0.15\height}{\begin{minipage}[b]{\textwidth}
      \includegraphics[width=\linewidth]{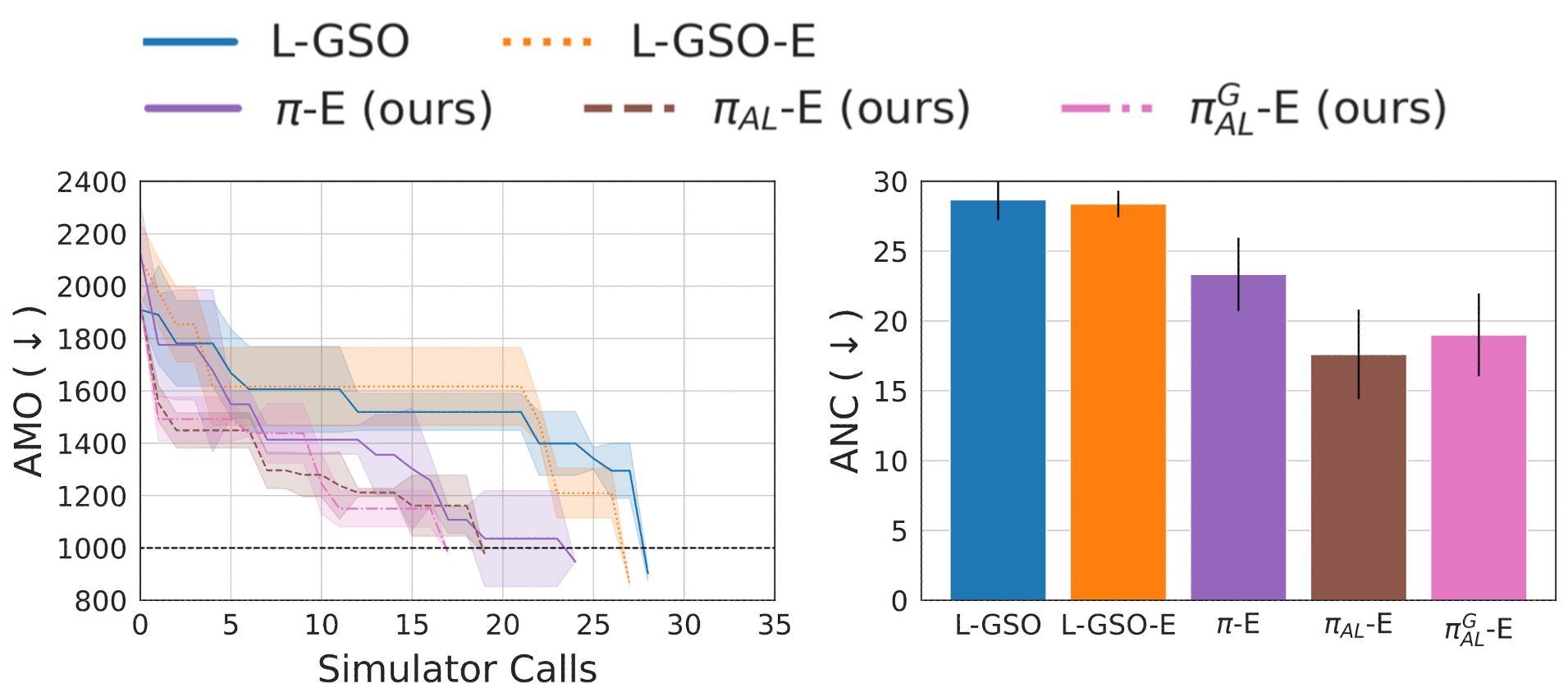}
      \caption{\textbf{Muon background reduction} results for fixed $\vx$-distribution.}
      \label{fig:ship_res}
    \end{minipage}}
    \vfill
    \begin{minipage}[b]{\textwidth}
      \caption{Wireless simulation results (office room) for a varying $\vx$-distribution.}
      \begin{center}
        \begin{tabularx}{\linewidth}{l|c|cc}
          \toprule
          & L-GSO-E & $\pi$-E & $\pi_{AL}$-E \\
          \cmidrule{2-4}
          ANC ($\downarrow$) & 23 & \textbf{7} & \textbf{8}  \\
          \bottomrule
        \end{tabularx}
        \label{tab:wireless_varying_x_dist}
      \end{center}
    \end{minipage}
  \end{minipage}
\end{figure}


\paragraph{Fixed $\vx$-distribution} We first focus on the experiments involving a fixed input distribution. These experiments will indicate how well the policy performs in simpler setting where the stochastic simulator has the same distribution during training and evaluation. Policy-based methods achieve the strongest overall performance on our benchmark functions, as shown in~\Figref{fig:benchmark_function_results_fixed_x_dist}. The $\pi$-E model scores best, especially on the average number of simulator calls (ANC) required to terminate an episode (bar plots in the figures) for almost all settings. In particular, using our trained policies, we observe a significant reduction in the number of required \textit{simulator calls} of up to $10\times$ with respect to L-GSO. Although $\pi_{AL}$-E outperforms all baselines, it slightly underperforms $\pi$-E. This may be because the trust region size $\epsilon$ for $\pi$-E and L-GSO is set to the optimal value reported in \citet{shirobokov2020lgso} in these benchmark experiments, which simplifies the problem relative to $\pi_{AL}$-E. Similarly, the AMO evaluations show that our policies outperform previous methods, with the single exception of the Nonlinear Submanifold Hump problem; here, our models are within error ranges of the best observed AMO. As noted in~\citep{shirobokov2020lgso}, the BOCK baseline struggles in solving the Rosenbrock problem (\Figref{fig:benchmark_function_results_fixed_x_dist}, middle row), likely due to the high curvature of the objective function under analysis. For our experiments on real-world black-box simulators, we use both variants of L-GSO as baseline methods, since they generally outperform the other baselines (and are on par in the worst case). Again, we see that policy-based methods achieve the strongest performance on both AMO and ANC. However, in contrast to the results of~\Figref{fig:benchmark_function_results_fixed_x_dist}, the error ranges of the policy methods overlap more closely (\Figref{fig:wireless_fixed_x_dist}). In the particle physics experiment, $\pi_{AL}$-E and $\pi_{AL}^G$-E perform better on average than $\pi$-E (\Figref{fig:ship_res}). This may suggest that learning to adapt the trust region size is advantageous primarily in more complex optimization problems.

\paragraph{Varying $\vx$-distribution} The reported AMO and ANC metrics do not include simulator calls incurred during policy training, so the previous comparisons do not account for the total number of simulation calls performed in order to obtain the optimized simulator parameter. Although the number of simulator calls we have used during policy training may still be heavily reduced (see, e.g., Appendix~\ref{sup:surrogate} and~\Figref{fig:three_hump_psi_pol_iter}), training a policy network is unlikely to be more efficient than direct optimization with, e.g., L-GSO on a single problem. Instead, the primary value of our policy-based methods is that a trained policy reduces the number of required simulator calls for downstream optimization problems; that is, training a policy amortizes downstream costs in settings where practitioners need to solve similar optimization problems many times. To operationalize this notion of `similar downstream optimization problem', we vary the distribution from which $\vx$ is sampled every optimization episode during training and evaluation. Thus, every episode corresponds to an optimization problem with a different input distribution (e.g., carrier frequency), as is typically encountered in real-world experiment design settings. The results on benchmark functions and wireless simulations in~\Figref{fig:benchmark_function_results_varying_x_dist} and Table~\ref{tab:wireless_varying_x_dist}, respectively, indicate that our policies are also effective for this setting. 
On both the Three Hump and Rosenbrock problem, we observe almost no reduction in AMO or ANC performance w.r.t. the fixed distribution setting. Note that $\pi_{AL}^G$-E outperforms $\pi_{AL}$-E here, potentially because the warm-started surrogate reduces the importance of a well-chosen $\epsilon$. On the Nonlinear Submanifold, $\pi$-E and $\pi_{AL}^G$-E performance is reduced but still on par with the best-performing alternative. Here, $\pi_{AL}$-E achieves the strongest performance, improving slightly over the fixed distribution setting. Given the computational cost of running real-world simulators (see \Appref{sup:exp_comp_res}), we report in Table~\ref{tab:wireless_varying_x_dist} a comparison between the strongest baseline, L-GSO-E, and two of our policies, where the policies outperform the baseline about $3\times$ on average. We leave expanding on these results to future work. In \Appref{sup:num_sims} we estimate efficiency gains using our policies in terms of simulator calls for the wireless and high energy physics settings, and discuss the break-even point between base L-GSO and our methods for wireless, which is at around 100 optimizations. These results suggest that our policies effectively amortize optimization costs. 
\section{Conclusion} \label{sec:conclusions}
We propose a novel method for minimizing the number of \textit{simulator calls} required to solve optimization problems involving black-box simulators using (local) surrogates. The core idea of our approach is to learn an active learning policy that controls \textit{when} the black-box simulator is used and \textit{how} to sample data to train the local surrogates. We describe three policy model variations and present experiments showing they outperform previous methods, including local surrogate methods~\citep{shirobokov2020lgso}, numerical differentiation, and Bayesian approaches~\citep{oh2018bock} in a variety of setups that include benchmark functions and real-world black-box simulators. In particular, we observe a significant reduction in the number of \textit{simulator calls} up to $10\times$. Our results suggest that local surrogate-based optimization of problems involving black-box forward processes benefits from the guidance of both simple policies and learned sampling strategies.



\bibliographystyle{plainnat}
\bibliography{references.bib}

\clearpage


\appendix

\section{Impact statement, limitations and future work}\label{sup:brd_imp_fut_w}

\subsection{Impact statement} \label{sup:impact}
This paper proposes a novel policy-based approach to guide local surrogate-based problem optimization with black-box simulators. We believe the potential societal consequences of our work are chiefly positive, as it has the potential to accelerate experiment design in various scientific domains, particularly those involving optimization procedures involving black-box, non-differentiable, forward processes. However, it is crucial to exercise caution and thoroughly comprehend the behavior of the models to obtain tangible benefits. Application of the models outside of their domain may lead to inefficiently designed real-life experiments, with all its associated costs.

\subsection{Limitations and future work} \label{sup:lim}
Gradient-based optimization may get stuck in local optima of the loss surface $\mathbb{E}_{p(\bm{y}|\bm{\psi}, \bm{x})} \left[ \mathcal{L}(\bm{y}) \right]$. Investigating whether introducing a policy into the optimization can help avoid such local minima is an interesting direction of future research. The Probabilistic Three Hump problem has no local minima but does contain a few flat regions, where gradient-based optimization is more challenging. Exploratory experiments have provided weak evidence that the policy may learn to avoid such regions.

Hyperparameter tuning has mostly involved reducing training variance through tuning the number of episodes used for a PPO iteration, as well as setting learning rates and the KL-threshold. Little effort has been spent optimizing the policy or surrogate architectures; we expect doing so to further improve performance. Surrogates (and thus policies) can likely be trained using fewer simulator datapoints per simulation call, which should be explored in future work. Additionally, it would be interesting to investigate whether the semi-local surrogate obtained from training $\pi_{AL}^G$-E can be used for simulator-less optimizations: i.e., not retraining the surrogate during optimization. 

More recent algorithms for policy training may offer advantages compared to \ac{PPO} with a value function critic, such as improved planning and off-policy learning for more data-efficient training \citep{haarnoja2018soft, neumann2023greedy}. 

When the policy decides how (with what data) the surrogate should be updated, it does so using information provided by the surrogate itself. This is an example of active learning \citep{settles2009al}, where the current instance of a task model (the surrogate) affects the data it sees in future training iterations. In particular, our policies are instances of \textit{learning active learning}, where a separate model (our policy) is trained to suggest the data that the task model should be trained on \citep{lal2015orig, lal2017, lal2017dqn, ravi2018meta, lal2018metarl, lal2018imit, lal2018uci, bakker2023lal}. Progress in the field of (learning) active learning may be applicable in our setting as well.

In this work, we studied the generalization of our policies to a varying distribution of input variables $\vx$. Another valuable setting is one in which the simulator optimization objective changes between runs, such as when doing sequential sensor placement. This is currently out of scope of our work, as properly integrating such dynamics into the policy would involve training it on such sequential optimization trajectories. However, there may be a reason to expect our policies to be somewhat robust to distribution shifts generally, namely that they only interact with the distribution shift through the surrogate uncertainty feature. Thus, if such distribution shifts do not strongly influence the relationship between surrogate uncertainty and good policy decisions, performance may be retained. Such a condition may, for instance, hold when the policy is implementing a threshold function based on surrogate uncertainty (e.g., uncertainty above a threshold leads to calling the simulator). We do not study this here, however.
\section{Policy and surrogate details} \label{sup:implementation}

\subsection{Policy} \label{sup:policy}


\begin{figure}[ht]
  \begin{subfigure}[t]{0.48\linewidth}
    \includegraphics[width=\linewidth]{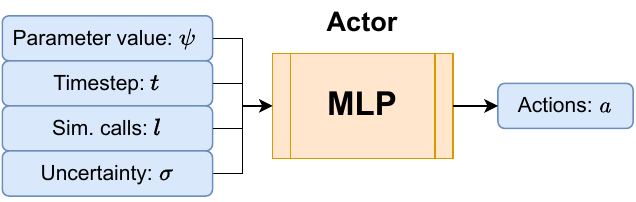}
    \caption{\textbf{Actor MLP}. The Actor output the actions: these always contain the decision $b$ to perform a simulator call or not and may optionally also contain a value $\epsilon$ used for surrogate training data sampling.}
    \label{fig:actor}
  \end{subfigure}
  \hfill
  \begin{subfigure}[t]{0.48\linewidth}
    \includegraphics[width=\linewidth]{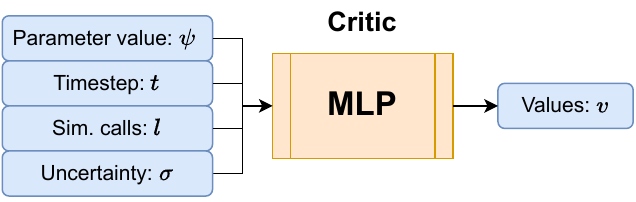}
    \caption{\textbf{Critic MLP}. The Critic outputs value function estimates used in the PPO objective.}
    \label{fig:critic}
  \end{subfigure}%
  \caption{Schematic for the Actor (left) and Critic (right) MLPs, which together form the policy. Both models take $(\bm{\psi}, t, l, \sigma)$ as input.}
\end{figure}

The policy $\pi_\theta$ is composed of two separate neural networks: an Actor and a Critic. Both networks are \ac{ReLU} \ac{MLP}s with a single hidden layer of 256 neurons, schematically depicted in~\Figref{fig:actor} and in~\Figref{fig:critic}. The input to both networks is the tuple: $(\bm{\psi}_t, t, l_t, \sigma_t)$, where $\bm{\psi}_t$ is the current parameter value (at timestep $t$), $l_t$ is the number of simulator calls already performed this episode, and $\sigma_t$ is the standard deviation over the average surrogate predictions in the ensemble.

The Actor outputs either one or three values. The first value is passed through a sigmoid activation and treated as a Bernoulli random variable, from which we sample $b$, representing the decision to perform a simulator call or not. If the policy outputs three values, the second and third values are treated as the mean and standard deviation of a lognormal distribution from which we sample $\epsilon$, the trust region size, for the current timestep. The standard deviation value is passed through a softplus activation function to ensure it is positive.

The Critic outputs a value-function estimate $V_\theta(s)$, where $\theta$ are policy parameters. We use this estimate to compute advantage estimates in PPO, as explained in detail in section~\ref{sup:ppo}. Since rewards have unity order of magnitude, we expect return values to be anywhere in $[-T, 0]$. To prevent scaling issues, we multiply the Critic output values by $T$ before using them for advantage estimation.

\paragraph{The $\pi_{AL}^{G}$ method} 
When training a policy for downstream optimization of many related black-box optimization problems, it may be helpful to train a global surrogate simultaneously for such a problem setting. Such a global surrogate might provide better gradients for problem optimization, especially if it has been jointly optimized with the policy. We have implemented the $\pi_{AL}^{G}$ method to test this. Here, the policy outputs both the decision to perform a simulator call and the trust-region size, $\epsilon$, just as in $\pi_{AL}$. However, the surrogate ensemble is ``warm-started'' from the previous training step every time a retraining decision is made. This results in a continuously optimized surrogate ensemble for the training trajectories. To prevent the surrogate from forgetting old experiences too quickly, we employ a replay buffer that undersamples data from earlier iterations geometrically. Specifically, when training the surrogate with trust-region $U_\epsilon^{\bm{\psi}}$, we include all data inside $U_\epsilon^{\bm{\psi}}$ for the current episode, half of the data inside $U_\epsilon^{\bm{\psi}}$ from the previous episode, a quarter of the data seen two episodes ago, and so on. 

\subsection{Surrogate}\label{sup:surrogate}
\begin{figure}[!ht]
  \centering
  \includegraphics[width=\linewidth]{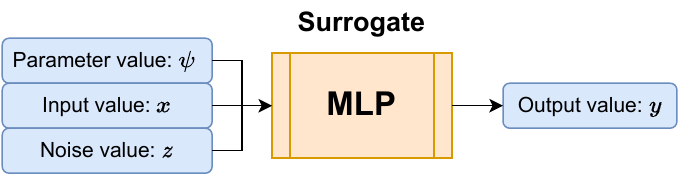}
  \caption{Schematic for the surrogate architecture. The surrogate is an MLP trained to mimic the simulator. It takes $(\bm{\psi}, \bm{x}, \bm{z})$ as input and outputs $\bm{y}$.}
  \label{fig:surrogate}
\end{figure}
Each surrogate model consists of a \ac{ReLU} \ac{MLP} with two hidden layers of 256 neurons that takes as input $(\bm{\psi}, \bm{x}, \bm{z})$ and outputs $\bm{y}$. $\bm{z}$ is sampled from a 100-dimensional diagonal unit Normal distribution. The surrogate architecture is schematically depicted in Figure~\ref{fig:surrogate}.

Surrogates are trained on data generated from $f_{\textrm{sim}}$. Following the approach outlined in \citet{shirobokov2020lgso}, we sample $M$ values $\bm{\psi}_j$ inside the box $U_\epsilon^{\bm{\psi}}$ around the current parameter value using an adapted Latin Hypercube sampling algorithm. For each of those $\bm{\psi}_j$, we then sample $N = 3 \cdot 10^3$ $x$-values. We use $M = 5$ for the Probabilistic Three Hump problem, $M = 16$ for the Rosenbrock problem, and $M=41$ for the Nonlinear Submanifold Hump problem. As in \citet{shirobokov2020lgso}, this means a single ``simulator call'' consists of $1.5 \cdot 10^4$ function evaluations for Probabilistic Three Hump, $4.8\cdot10^4$ for Rosenbrock, and $1.23 \cdot 10^5$ for the Nonlinear Submanifold Hump. In the Wireless setting, we instead use $N = 500$ and $M=6$, resulting in only $3.0\cdot10^3$ function evaluations per simulator call. This suggests that, in our setting, the number of function evaluations (forward passes) per simulator call in the benchmark experiments can be heavily reduced from the defaults mentioned in \citet{shirobokov2020lgso}, but we do not experiment with this, as these evaluations can all be done in parallel at low marginal computational cost in a single (expensive) ``simulator call''.

The setup for the Physics setting is slightly different. Following \citet{shirobokov2020lgso}, we employ a fixed dataset of $1.50\cdot10^5$ muon events, all sampled from a particular muon input ($\vx$ distribution). We split this into two datasets; $10^5$ for training and $5.0\cdot10^4$ for evaluation. Every time the simulator is called, we sample $5.0\cdot10^4$ $\vx$ from the respective dataset. Thus, every optimization trajectory uses at most $10^5$ samples, but events are used many times per optimization.

To train the surrogates, we use the Adam optimizer with MSE loss for two epochs with a learning rate of $10^{-3}$ and a batch size of $512$. Each \emph{surrogate ensemble} comprises \emph{three} surrogates, each trained on identical data but a different random seed. The uncertainty feature $\sigma$ is computed using mean predictions of each individual surrogate in the ensemble. Specifically, we compute the prediction mean per surrogate on $D$ samples as $\bm{\bar{y}} = \frac{1}{D} \, \sum_{i=1}^D \left[ f_\phi(\bm{\psi}, \bm{x}_i, \bm{z}_i) \right]$, and construct $\sigma$ as the standard deviation over these mean predictions. We use $D=100$ in all our experiments.

\paragraph{Surrogate reliability:} Surrogate reliability in locally mimicking the simulator was assessed by inspection of the MSE loss, manually checking that surrogate predictions matched simulator predictions, and visualizing the local gradient field of the surrogate for the 2D Three Hump Problem and checking that it (locally) matched the changes in objective value as a function of $\psi$. The policy itself assesses surrogate reliability through the ensemble uncertainty feature $\sigma$, which captures underspecification of the surrogates given the current simulator data.

\subsection{Full observability of the MDP} \label{sup:pomdp}
Because the state of our reinforcement learning framework consists of the fully observed variables $(\bm{\psi}_t, t, l_t, \sigma_t)$, we have formulated it as an MDP rather than as a POMDP (Partially Observable MDP). Concretely, our method can be applied if the parameter setting $\bm{\psi}$ of a simulator is known at all times, since $t$ and $l_t$ increment based on policy decisions and $\sigma_t$ is generated using separate surrogate models. Training these surrogate models requires $(\bm{\psi}, \bm{x}, \bm{z})$, where $\bm{x}$ and $\bm{x}$ are user-generated and $\bm{y}$ is observed simulator output. If $\bm{\psi}$ is not observed, then backpropagation through the surrogate w.r.t. $\bm{\psi}$ is not possible, and our method is not applicable. However, note that -- even in black-box optimisation settings -- the simulator parameter settings $\bm{\psi}$ are generally input values specified by the user, and thus observed.
\section{Simulator details}\label{sup:exp_details}

\subsection{Benchmark Functions} \label{sup:sim}
Our tests with benchmark functions employ a probabilistic version of three benchmark functions from the optimization literature: Probabilistic Three Hump, Rosenbrock, and Nonlinear Submanifold Hump. The former is a two-dimensional problem that lends itself well to visualization. The $N$-dimensional Rosenbrock (with $N=10$) and the Nonlinear Submanifold Hump problems are used to test our method on higher-dimensional settings. 

\paragraph{Probabilistic Three Hump Problem} 
The goal is to find the 2-dimensional $\bm{\psi}$ that optimizes:\footnote{Here the upper bound of $x_1$ and lower bound of $x_2$ are switched compared to the notation in Equation (3) of \citep{shirobokov2020lgso}. These bounds match the official implementation of L-GSO as of August 2023. 
}
\begin{gather}
    \begin{align}
        \bm{\psi}^* &= \argmin_{\bm{\psi}} \, \mathbb{E} \left[ \mathcal{L}(y) \right] \nonumber \\
        &= \argmin_{\bm{\psi}} \, \mathbb{E} \left[ \sigma(y - 10) - \sigma(y) \right], \nonumber \text{s.t.}
    \end{align} \\
    y \sim \mathcal{N}(y|\mu_i, 1), \, i \in \{1, 2\}, \; \mu_i \sim \mathcal{N}(x_i h(\bm{\psi}), 1), \; \nonumber \\
    x_1 \sim U[-2, 2], \; x_2 \sim U[0, 5], \nonumber \\ 
    P(i=1) = \frac{\psi_1}{||\bm{\psi}||_2} = 1 - P(i = 2), \; \nonumber\\
    h(\bm{\psi}) = 2 \psi_1^2 - 1.05 \psi_1^4 + \psi_1^6 / 6 + \psi_1 \psi_2 + \psi_2^2.
\end{gather}
We consider an episode terminated when $\mathbb{E} \left[ \mathcal{L}(y) \right] = \frac{1}{N} \, \sum_{i=1}^N \mathcal{L}(f_{\textrm{sim}}(\bm{\psi}, \bm{x}_i)) \leq \tau = -0.8$, which we evaluate after every optimization step using $N=10^4$ samples. Following \citep{shirobokov2020lgso}, we use $\epsilon=0.5$ as the trust-region size. The optimization is initialized at $\bm{\psi}_0 = [2.0, 0.0]$; this is a symmetry point in the Three Hump function such that optimization with stochastic gradients can fall into either of the two wells around the two minima of the function. Such a procedure requires our methods to learn good paths to both optima, making the task more interesting. In principle, optimization could be initialized at any $\bm{\psi}_0$.

\paragraph{Rosenbrock Problem} 
The goal for this problem is to find the 10-dimensional $\bm{\psi}$ that optimizes: 
\begin{gather}
\bm{\psi}^* = \argmin_{\bm{\psi}} \, \mathbb{E} \left[ \mathcal{L}(y) \right] = \argmin_{\bm{\psi}} \, \mathbb{E} \left[ y \right] \nonumber \\
y \sim \mathcal{N}\Bigg(y;\ \sum^{n-1}_{i=1} \left[ (\psi_{i+1} - \psi_i^2)^2 + (\psi_i - 1)^2 \right] + x,\ 1 \Bigg), \nonumber \\
x\sim \mathcal{N}(x;\mu,1);\ \mu \sim \mathrm{U} \left[ -10,10 \right].
\end{gather}
We consider an episode terminated when $\mathbb{E} \left[ \mathcal{L}(y) \right] = \frac{1}{N} \, \sum_{i=1}^N \mathcal{L}(f_{\textrm{sim}}(\bm{\psi}, \bm{x}_i)) \leq \tau = 3.0$, which we evaluate after every optimization step using $N=10^4$ samples. Following \citep{shirobokov2020lgso}, we use $\epsilon=0.2$ as the trust-region size and $\bm{\psi}_0 = [\overrightarrow{2.0}] \in \mathbb{R}^{10}$ to initialize the optimization.

\paragraph{Nonlinear Submanifold Hump Problem} 
In this problem, we seek to find the optimal parameters vector $\vpsi$ in $\mathbb{R}^{40}$ by utilizing a non-linear submanifold embedding represented by $\hat{\vpsi}=\mB \mathrm{tanh}(\mA\vpsi)$, where $\mA \in \mathbb{R}^{16 \times 40}$ and $\mB \in \mathbb{R}^{2 \times 16}$.
To achieve this, we use $\hat{\vpsi}$ in place of $\vpsi$ in the Probabilistic Three Hump problem definition. Here too we use similar settings as in~\citep{shirobokov2020lgso}: the orthogonal matrices $\mA$ and $\mB$ are generated via a \textit{QR}-decomposition of a random matrix sampled from the normal distribution; we use $\epsilon=0.5$ as the trust-region size and initialize the optimization at $\bm{\psi}_0 = [2.0 , \overrightarrow{0.0}] \in \mathbb{R}^{40}$.

\subsection{Real-world Simulators} \label{sup:real_world_sim}

\paragraph{Wireless Communication: Indoor Transmitting Antenna Placement}
The goal in this scenario is to find an optimal \textit{transmit} antenna location $\vpsi$ that maximizes the signal strength over multiple \textit{receiving} antenna locations $\vx \sim q(\vx)$. We run wireless simulations using Matlab's Antenna Toolbox \cite{matlab2022}, by evaluating the received signal strength (\texttt{sigstrength} function).
The simulations are run in two 3d scenes (`\texttt{conferenceroom}' and `\texttt{office}'), both of which are available by default, and we additionally let Matlab automatically determine the surface materials.
We use the `raytracing` propagation model with a maximum of two reflections and by disabling diffraction.
The objective is to find a transmit antenna location $\vpsi$ to maximize the received signal strength over locations $\vx \sim q(\vx)$.
We constrain the locations in a 3D volume spanning the entire XY area of the two scenes: 3$\times$3m in \texttt{conferenceroom} and 8$\times$5m in \texttt{office}.
The transmit elevations $\psi$ are constrained between 2.2-2.5m and 3.0-3.2m per scene, and the receive locations between 1.3-1.5m (identical for both scenes).
The end-objective is to identify a transmit location $\vpsi$ such that the median receive signal strength is maximized over a uniform distribution of receive antenna locations $q(\vx)$. For experiments with varying input distribution, we vary the carrier frequency $x_{f_c}$, which may be considered part of the simulator input $\vx = (x_1, x_2, x_3, x_4)$, where $x_{1:3}$ are antenna location coordinates and $x_4 = x_{f_c}$.
We use $\epsilon=0.4$ as the trust-region size and initialize the optimization at $\bm{\psi}_0 = (-1.35,-1.35,2.35)$ for the conference and $\bm{\psi}_0 = (3.0, 7.0, 3.05)$ for the office setting (xyz coordinates in meters).

\paragraph{Particle Physics: Muon Background Reduction}
In the high energy physics (HEP) experiments, we deal with the problem of reducing the number of unwanted events in the so-called \textit{sensitive region} of the detector. Specifically, we want to reduce the number of events from muons, which are notoriously a major contribution to the background noise in HEP detectors. 
We use the Geant4~\citep{agostinelli2003geant} and FairRoot~\citep{alturany2012} toolkits to simulate such events. The input $\vx \in\mathbb{R}^7 \sim q(\vx)$ comprises kinematic properties of the incoming muons, such as the momentum, the azimuthal and polar angles with respect to the incoming $z$-axis, as well as the charge, and the 3D coordinates of the particles' starting point. The parameters to be optimized, $\bm{\psi}\in\mathbb{R}^{42}$, describe the geometric properties of the six magnets that make up the SHiP muon shield (see ~\Figref{fig:ship_pic}). The available dataset has a size of $\mathcal{O}(10^5)$ that we split into training and validation sets. We use $\epsilon=10$ for the trust region size. 
The loss function used in the physics experiments encourages the muons to sweep away along the longitudinal axis of the detector in order not to hit the sensitive region placed behind the muon shield. 
We refer to Appendix \ref{sup:exp_comp_res} for a discussion of computation resources.

\subsection{Varying Input Distribution} \label{sup:param_input}
The goal of using a varying $\vx$-distribution is to emulate scenarios in which practitioners need to run the same type of simulator optimization multiple times. For example, as we reported in the main corpus of the paper, this is typically the case when optimizing antenna positioning in an indoor environment for which the same simulation needs to be run multiple times across different frequency bands. Here we discuss how we implemented this setting in our experiments.
\paragraph{Benchmark functions}
We randomly sample the upper, $u_x$, and lower, $l_x$, bounds of the $\vx$ distribution for each optimization run. Specifically, we assume the boundaries were normally distributed and sampled their value at each run. Once we obtain the bounds, we use them to construct a uniform distribution, $q(\vx) = U_{l_x, u_x}$, to sample $\vx$ values. In all cases, the Normal distributions on the bounds are centered on the value used in the fixed $\vx$-distribution setting; e.g., for the Three Hump problem, we sample $(l_{x_1}, u_{x_1})$ from Normals centered at -2 and 2 respectively, since $x_1 \sim U_{-2, 2}$ originally. The Normals have standard deviation 0.5 for $x_1$ and 1 for $x_2$ in Three Hump and Submanifold Three Hump, and standard deviation 2 for Rosenbrock.
\paragraph{Wireless Communication: Indoor Transmitting Antenna
Placement} To simulate a varying input distribution to the simulator, we randomly sample the carrier frequency $x_{f_c}$ passed to the ray tracer for each optimization episode. The main challenge in such a setup comes from the different behaviors of electromagnetic waves at different wavelengths. Specifically, we sampled frequencies uniformly in the $[4, 8]$ GHz range.

\subsection{Objective Landscape and Optimization Trajectory}\label{sup:loss_traj_land}
Experiments with low-dimensional functions, such as the Probabilistic Three Hump problem ($\vpsi \in \mathbb{R}^2$), allow us to easily visualize optimization trajectories to gain insights into the models behavior.

\begin{figure*}[!ht]
  \centering
  \includegraphics[width=\linewidth]{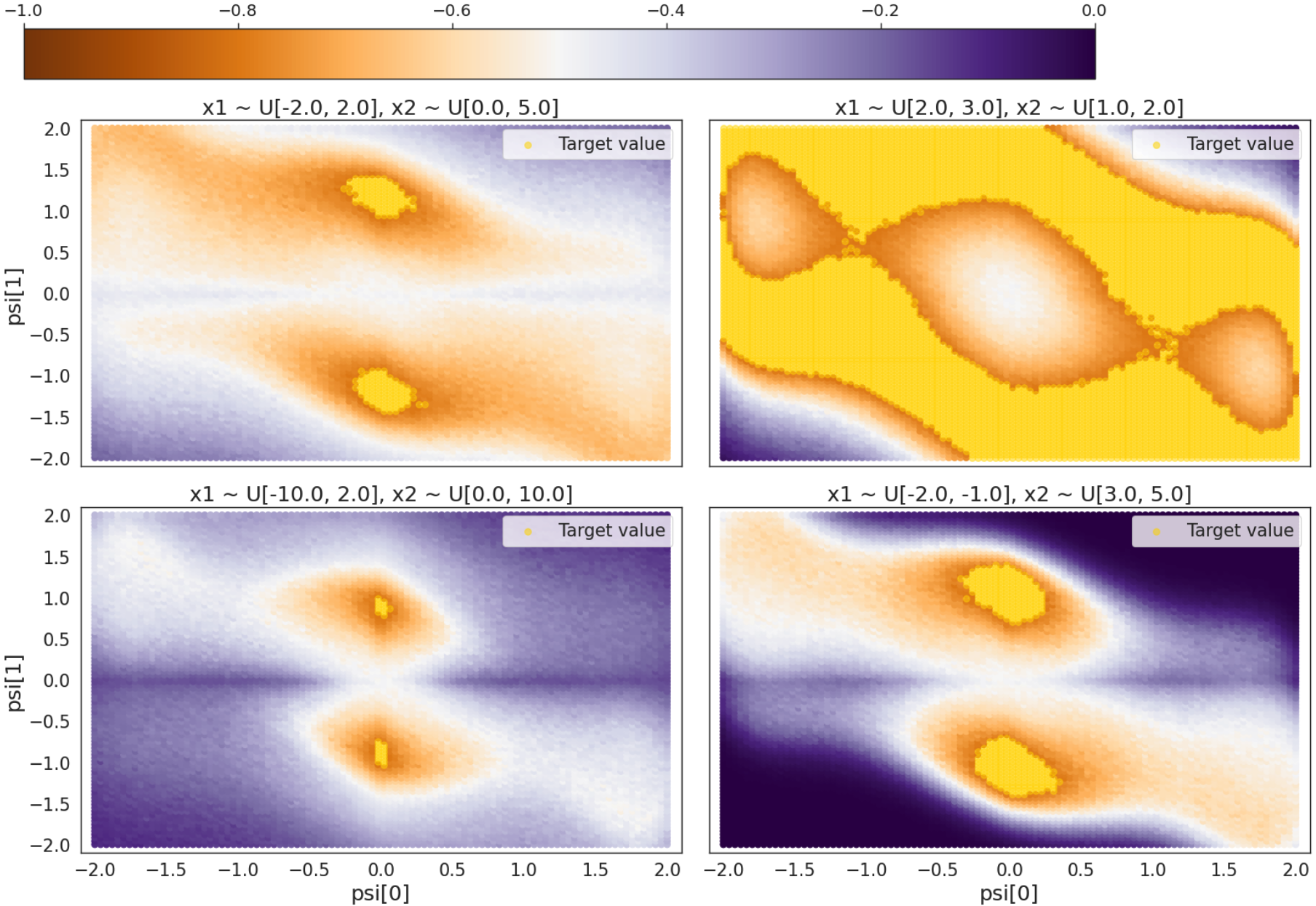}
  \caption{Loss landscape for the Probabilistic Three Hump Problem for different bounds on the $x$-distribution. Shown is $\frac{1}{n} \, \sum_{i=0}^{n-1} \mathcal{L}(f_s(\bm{\psi}, \bm{x}_i))$ for a grid of $\bm{\psi}$ values ($n = 100$). The yellow region denotes $\bm{\psi}$ values that lead to termination.}
  \label{fig:three_hump_psi_surface}
\end{figure*}

As mentioned in the main text, practitioners in many scientific fields may need to solve a set of related black-box optimization problems that can become costly if each optimization process has to begin \textit{ab initio}. Therefore, we investigated the robustness of the policy trained on a given set-up, that is, input $\vx$-distributions, and then tested on different ones. To mimic such a scenario, we consider a parameterized input $\vx$-distribution. In real-world experiments, this variation could correspond to different properties of the input data used to run the simulations. We already report the results on such tests in~\Secref{sec:exp_res}. In~\Figref{fig:three_hump_psi_surface}, we show the optimization landscape for different $\vx$-distributions of the Probabilistic Three Hump problem. 
We note that, although the minima are generally in the same location of $\vpsi$-space, the landscape dramatically changes from one distribution to another. 

\subsection{Metrics} \label{sup:metrics}
As we mentioned in~\Secref{sec:exp_res}, we use two metrics to compare our models against the baselines: the Average Minimum of the Objective function (AMO) for a specific budget of simulator calls and the Average Number of simulator Calls (ANC) required to terminate an episode. We now delve deeper into both of them. The meaning of the latter is quite straightforward. We consider the average number of simulator calls to solve the problem. We compute the average across evaluation episodes and random seeds. In contrast, the AMO is slightly less intuitive to interpret. One might question whether the value of the ANC should align with the maximum value on the $x$-axes for the AMO. In other words, assuming that for a given model, the ANC is equal to, e.g. 10, \emph{should one expect that at a value $x=10$, the AMO will be equal to the termination value?} 
Generally speaking, the answer is \emph{no}. To explain why that is the case, we can report the following example. Let us assume that, for a given model, we have the following three episodes, each characterized by a specific length and value of the objective function at each simulator call:
\begin{itemize}
    \item Episode 1: [20, 12, 7, 5, 3, 1]
    \item Episode 2: [18, 6, 1]
    \item Episode 3: [15, 5, 1]
\end{itemize}
We assumed the target value, $\tau$, to be 1. For simplicity, we used integers for the objective value. As we can see from the example, we have $\mathrm{ANC} = 4$. Now, if we examine the AMO for $x=4$, we find that it is equal to 5 since only the first episode contributes to it, which is greater than $\tau$. Therefore, one cannot directly map the $x$-axis from the AMO to the $y$-axis of the ANC. Such a one-to-one mapping would exist only when all episodes always require the same number of simulator calls, which is not the case. We hope that our explanation has clarified the interpretation of the results we reported in the main corpus of the paper. 
\section{Experiment details}\label{sup:training_details}

\subsection{Training} \label{sup:ppo}

\begin{table*}
    \caption{Hyperparameters used for RL training}
    \begin{center}
    
    \begin{tabularx}{\textwidth}{p{0.65\linewidth}|l|p{0.25\linewidth}|l}
    
    \toprule

    \multicolumn{2}{X}{Episodes}  \\
    \midrule
    Maximum number of timesteps $T$ & 1000 \\
    Maximum number of simulator calls $L$ (High Energy Physics) & 30 \\
    Maximum number of simulator calls $L$ (rest) & 50 \\
    Episodes per PPO iteration (High Energy Physics) & 5 \\
    Episodes per PPO iteration (Submanifold Hump \& Wireless) & 10 \\
    Episodes per PPO iteration (ThreeHump \& Rosenbrock) & 16 \\
    Number of episodes for evaluation (High Energy Physics) & 5 \\
    Number of episodes for evaluation (Wireless) & 10 \\
    Number of episodes for evaluation (rest) & 32 \\
    
    \midrule
    \multicolumn{2}{X}{PPO} \\
    \midrule
    Discount factor $\gamma$ & 1.0 \\
    GAE $\lambda$ & 0.95 \\
    Clip value & 0.2 \\
    Maximum number of actor updates per iteration & 20 \\
    KL threshold (simulator call action) & $3\cdot 10^{-3}$ \\
    KL threshold (trust-region size action) & $10^{-2}$ \\
    Actor learning rate (Adam) & $3\cdot 10^{-4}$ \\
    Maximum number of critic updates per iteration & 10 \\
    Critic threshold (MSE) & $30.0$ \\ 
    Critic learning rate (Adam) & $10^{-4}$ \\

    \bottomrule
    
    \end{tabularx}
    
    \label{tab:rl_hyp}
    \end{center}
    \vspace{-0.5cm}
\end{table*}

We train our policy in an episodic manner by accumulating sequential optimization episodes and updating the policy using \ac{PPO} \citep{schulman2017ppo} with \ac{GAE} advantages \citep{schulman2016gae} (discount factor $\gamma=1.0$, GAE $\lambda=0.95$). Episodes terminate once any of the following conditions is met: A) the target value for the loss, $\tau$, has been reached, B) the number of timesteps $T=1000$ has been reached, or C) the number of simulation calls $L=50$ has been reached ($L=30$ for the particle physics settings, instead). For every training iteration, before doing PPO updates, we accumulate: 10 episodes for the Nonlinear Sub. Hump and 16 episodes for the Rosenbrock and Prob. Three Hump problems, 10 episodes for the wireless simulations and 5 for the high energy physics experiments. The different choices in the number of episodes to accumulate are mainly dictated by the time required to complete one episode.

We use the PPO-clip objective (with clip value $0.2$) on full trajectories (all episodes and timesteps are processed in one batch) with no entropy regularization to perform Actor updates. We perform multiple Actor updates with the same experience until either the empirical KL-divergence between the old and new policy reaches a threshold ($3 \cdot 10^{-3}$ for simulator-call decision actions, $10^{-2}$ for trust-region size $\epsilon$ actions), or 20 updates have been performed. In practice, we rarely perform the full 20 updates. Updates use the Adam optimizer with learning rate $3 \cdot 10^{-4}$.

Similarly, we perform multiple Critic updates using the Mean-Squared Error (MSE) between the estimated values $V_\theta(s_t)$ and the observed return (sum of rewards, as $\gamma=1.0$) $R_t$ at every timestep. We keep updating until either $\text{MSE} \leq 30.0$ or ten updates have been done. This approach helps the critic learn quickly initially and after seeing surprising episodes but prevents it from over-updating on similar experiences (as MSE will be low for those iterations). Updates use Adam with learning rate $10^{-4}$. See~\Algref{alg:active_train} for the training pseudo-code and~\Algref{alg:active_inf} for the evaluation procedure pseudo-code.

\begin{figure}[!t]
  \centering
  \includegraphics[width=\linewidth]{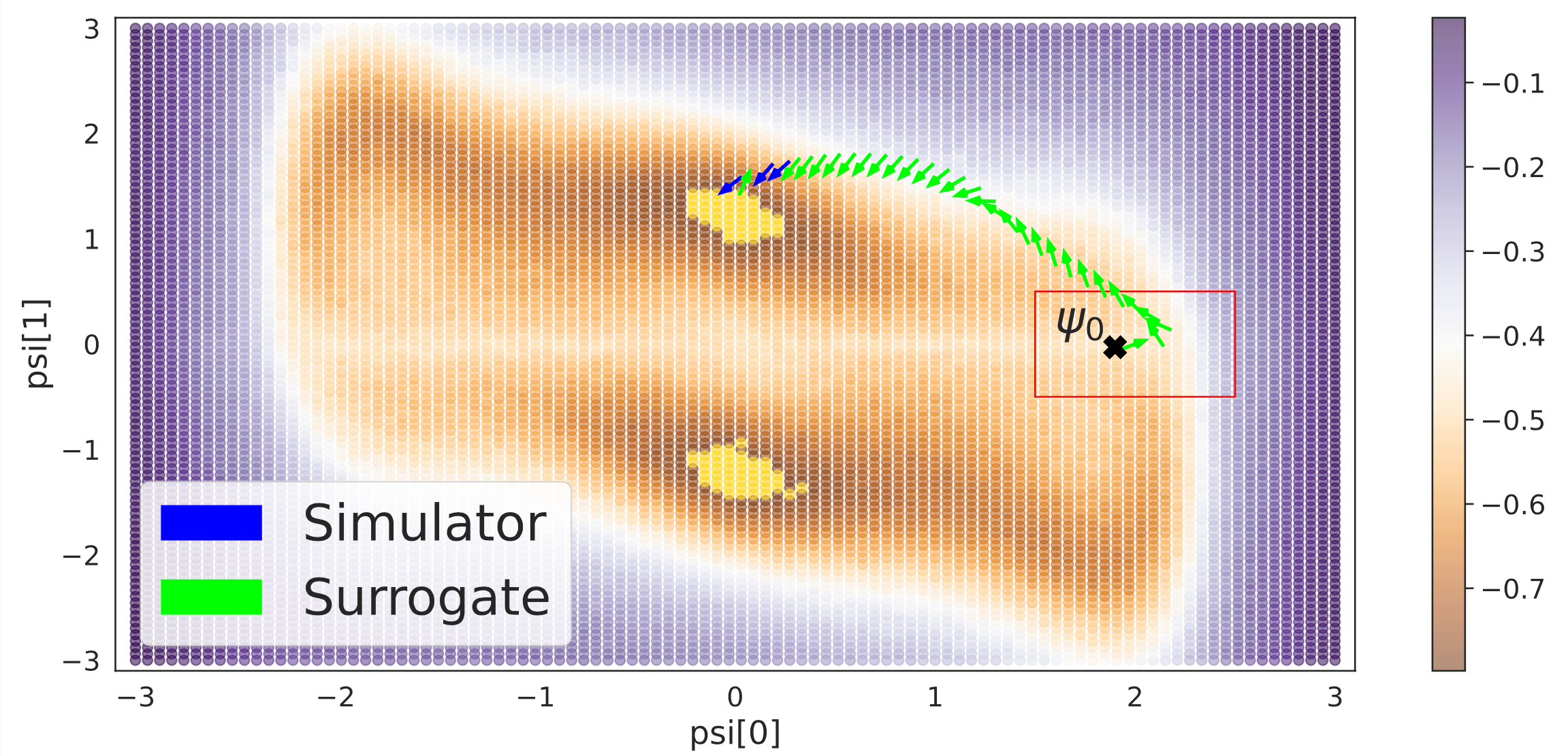}
  \caption{\textbf{Loss landscape and learned optimization trajectory for the 2D Three Hump problem}. The yellow region denotes $\bm{\psi}$ values that lead to termination. The $\epsilon=0.5$ region around $\bm{\psi}_0$ (black cross) is visualized as the red box. 
  Light green and blue arrows respectively represent gradients from a previously trained surrogate or after a simulator call.}
  \label{fig:main_three_hump_psi_opt_traj}
\end{figure}

To assess the performance of our models, we run 32 evaluation episodes for the benchmark functions and 20 and 5 evaluation episodes for the wireless and physics experiments, respectively. Moreover, we consider three random seeds for L-GSO and policy models, while we used ten random seeds for the BOCK and Num. Diff. baselines.

\begin{algorithm*}
    \caption{Training the (active learning) policy.} \label{alg:active_train}
    \KwData{Simulator $f_{\mathrm{sim}}(\bm{\psi}, \bm{x)}$; surrogate $f_\phi(\bm{\psi}, \bm{x})$; objective function $\mathcal{L}$; policy $\pi_\theta$; number $N$ of $\bm{\psi}$ to sample when training the surrogate; number $M$ of of $\bm{x}$ to sample for each $\bm{\psi}$; distributions $Q(q)$ over distributions $q(\bm{x})$ to sample $\bm{x}$ from; initial value $\bm{\psi}_0$; target function value $\tau$; number $T$ of timesteps to run each simulation for (episode-length); maximum number of simulator calls $L$; $\bm{\psi}$ optimiser $\textsc{optim}_\psi$ with learning rate $\lambda$; number of policy training iterations $K$; number of episodes to accumulate for a PPO step $G$; policy optimiser $\textsc{optim}_\pi$; reward function $\mathcal{R}$; experience buffer $B$; discount factor $\gamma$.}
    \For{$k \in (1, ..., K)$}{
        Empty experience buffer $B$. \\ 
        \For{$\_ \in (1, ..., G)$}{
            Initialise number of simulator calls done: $l_t \leftarrow 0$. \\
            Set return: $R \leftarrow 0$. \\
            Sample $\bm{x}$-distribution $q \sim Q$. \\
            \For{$t \in (1, ..., T)$}{
                Sample $\bm{x} \sim q(x)$. \\
                Obtain uncertainty $\sigma_t$ from the ensemble surrogate $f_\phi(\bm{\psi}_t, \bm{x})$. \\
                Construct state: $s \leftarrow (\bm{\psi_t}, t, l_t, \sigma_t)$. \\
                Obtain action: $a = (\texttt{do\_retrain}, \; \texttt{trust\_region\_size}) \leftarrow \pi_\theta(s)$. \\
                \If{\texttt{do\_retrain}}{
                    Obtain $N$ samples $\bm{\psi}_n$ from trust region with size $\texttt{trust\_region\_size}$. \\
                    Obtain $M$ samples $\bm{x}_m \sim q(\vx)$ for each of these $\bm{\psi}_n$. \\
                    Combine into dataset $\{ \bm{\psi}, \{ \bm{x} \}^M \}^N$ and optionally filter or include data from previous timesteps. \\
                    Retrain surrogate: $f_\phi$ on this dataset. \\
                    Increment number of simulator calls: $l_t \leftarrow l_t + 1$. \\
                }
                Obtain surrogate gradients: $\bm{g}_t \leftarrow \nabla_\psi \mathcal{L}(f_\phi(\bm{\psi}, \bm{x}) )|_{\bm{\psi}_t}$. \\
                Do optimisation step: $\bm{\psi}_{t+1} \leftarrow \textsc{optim}_\psi(\bm{\psi}_t, \bm{g}_t, \lambda)$. \\
                $\texttt{terminated} \leftarrow \mathbb{E} \left[ \mathcal{L}( f_{\mathrm{sim}}(\bm{\psi}_t, \bm{x} ) ) \right] \leq \tau$ \\
                Obtain reward: $r \leftarrow \mathcal{R}(s, a, \bm{\psi}_{t+1})$. \\
                Store $(s, a, r)$ and any other relevant information in buffer $B$. \\
                \If{\texttt{terminated}}{
                    \texttt{break}
                }
                \If{$l$ equals $L$}{
                    \texttt{break}
                }
            }
        }
        Update policy $\pi_\theta \leftarrow \textsc{optim}(\pi_\theta, B, \gamma)$.
    }
\end{algorithm*}

\begin{algorithm}
    \caption{Inference with the (active learning) policy.} \label{alg:active_inf}
    \KwData{Simulator $f_s(\bm{\psi}, \bm{x)}$; surrogate $f_\phi(\bm{\psi}, \bm{x})$; objective function $\mathcal{L}$; trained policy $\pi_\theta$; number $N$ of $\bm{\psi}$ to sample when training the surrogate; number $M$ of of $\bm{x}$ to sample for each $\bm{\psi}$; distributions $q(\bm{x})$ to sample $\bm{x}$ from; initial value $\bm{\psi}_0$; target objective value $\tau$; number $T$ of timesteps to run each simulation for (episode-length); maximum number of simulator calls $L$; $\bm{\psi}$ optimizer $\textsc{optim}_\psi$ with learning rate $\lambda$.}
    \For{$t \in (1, ..., T)$}{
        Initialise number of simulator calls done: $l_t \leftarrow 0$. \\
        Sample $\bm{x} \sim q(x)$. \\
        Obtain uncertainty $\sigma_t$ from the ensemble surrogate $f_\phi(\bm{\psi}_t, \bm{x})$. \\
        Construct state: $s \leftarrow (\bm{\psi_t}, t, l_t, \sigma_t)$. \\
        Obtain action: $a = (\texttt{do\_retrain}, \; \epsilon=\texttt{trust\_region\_size}) \leftarrow \pi_\theta(s)$. \\
        \If{\texttt{do\_retrain}}{
            Obtain $N$ samples $\bm{\psi}_n$ from trust region with size $\epsilon$. \\
            Obtain $M$ samples $\bm{x}_m$ for each of these $\bm{\psi}_n$. \\
            Combine into dataset $\{ \bm{\psi}, \{ \bm{x} \}^M \}^N$ \Comment{filter or include data from previous timesteps.} 
            Retrain surrogate: $f_\phi$ on this dataset. \\
            Increment number of simulator calls: $l_t \leftarrow l_t + 1$.
        }
        Obtain surrogate gradients: $\bm{g}_t \leftarrow \nabla_\psi \mathcal{L}(f_\phi(\bm{\psi}, \bm{x}))|_{\bm{\psi}_t}$. \\
        Do optimization step: $\bm{\psi}_{t+1} \leftarrow \textsc{optim}_\psi(\bm{\psi}_t, \bm{g}_t, \lambda)$. \\
        $\texttt{terminated} \leftarrow \mathbb{E} \left[ \mathcal{L}( f_s(\bm{\psi}_t, \bm{x} ) ) \right] \leq \tau$ \\
        \If{\texttt{terminated}}{
            \texttt{break}
        }
        \If{$l$ equals $L$}{
            \texttt{break}
        }
    }
\end{algorithm}

\begin{figure*}[!ht]
  \centering
  \includegraphics[width=0.87\linewidth]{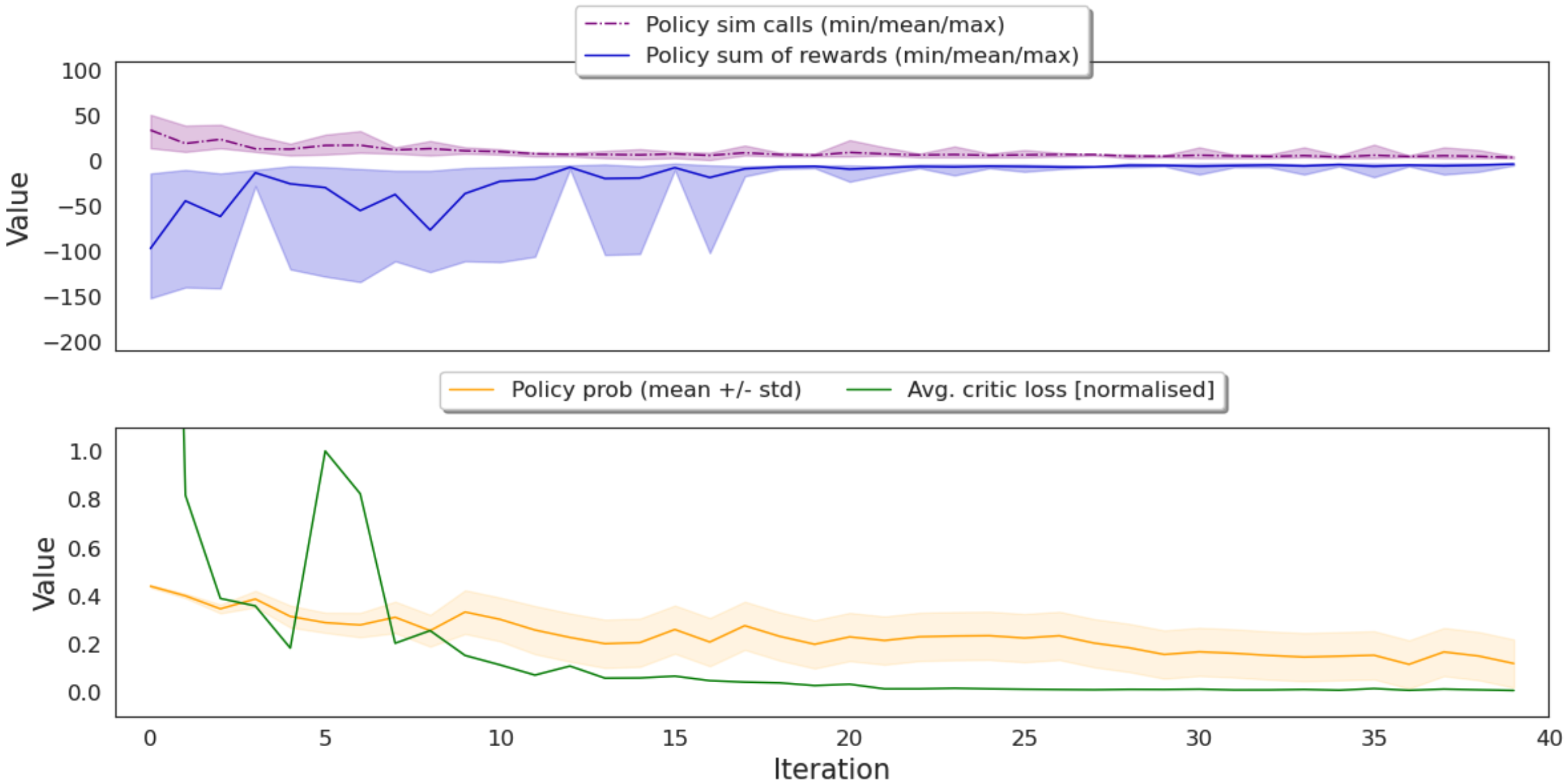}
  \caption{\textbf{Top}: Average number of simulator calls (purple curve) and average sum of rewards (blue curve) after each PPO iteration. Values are averaged across evaluation episodes. The upper and lower bound for the shadowed areas represent the max. and min. for each of the two mentioned metrics, respectively. \textbf{Bottom}: Average value of the probability of calling the black-box simulator (orange curve) and average critic loss (green curve). The shaded region represents the standard deviation. Values are averages across evaluation episodes and time steps within an episode for each PPO iteration. Note that the episodic return stabilizes before the 20th PPO iteration.}
  \label{fig:three_hump_psi_pol_iter}
\end{figure*}

\Figref{fig:three_hump_psi_pol_iter} shows that the policy is actually able to learn \textit{when} to call the simulator. Initially, during the first stages of training, the policy generates completely random actions, resulting in an average probability of calling the simulator close to $0.5$ (bottom plot in \Figref{fig:three_hump_psi_pol_iter}). However, as training progresses, this probability gradually decreases, leading to a reduction in the number of simulator calls (top plot in \Figref{fig:three_hump_psi_pol_iter}). Furthermore, the standard deviation of this probability -- computed over timesteps and episodes -- increases, which suggests that the policy is learning to differentiate when low and high probabilities are appropriate.

\subsection{Reward Design}\label{sup:rew_design}
The reward function is chosen to incentivize the policy to reduce the number of simulator calls. This is achieved by giving a reward of -1 every time the policy opts to call the simulator, contrasting with a reward of 0 when it does not. However, with this reward function the policy could achieve maximum return (of zero) by never calling the simulator even if this leads to non-terminating episodes. An extra term is required to make any non-terminating episodes worse than any terminating one. Since the minimum return is $-L$, corresponding to the maximum number of simulator calls for an episode, that is achieved by setting a reward penalty of $-(L-l_t)-1$ whenever the episode ends for reasons other than reaching the target value $\tau$: if the simulator call budget has been exhausted, then $l_t = L$ and the penalty is $-1$; if the timestep budget has been exhausted, then we have accumulated $-l_t$ return already. In both cases, adding this penalty leads to a total return of $-L-1 < -L$.

\subsection{Termination Value} \label{sup:term_val}
Termination values $\tau$ for the Probabilistic Three Hump and Rosenbrock problems are chosen to trade-off episode length and optimization precision. Selecting values very close to the exact minimum value of the objective function $\mathcal{L}$ leads to extremely long episodes, due to the stochastic nature of the optimization process. Moreover, varying the distribution of the $\vx$ variables changes the (expected) objective value minimum, such that choosing a too low value for $\tau$ leads to episodes that cannot terminate even in theory. Computing the minimum of $\mathcal{L}$ on the fly for the various $\vx$-distributions is not trivial, and so we opted for a $\tau$ that generally suffices for good performance across parameterizations of a given problem. These values are chosen by manually inspecting L-GSO runs. 

\subsection{Compute Resources}\label{sup:exp_comp_res}
Performing a single optimization trajectory for the benchmark functions does not require a significant amount of computational resources and can be conducted using any commercially available NVIDIA GPU. A single optimization can be easily fitted on a single GPU.
On the contrary, both wireless and physics experiments require extensive computing resources for running simulations. Especially for the physics setting, although it is still feasible to run the entire optimization on a single machine, simulation takes a considerable amount of time when simulating thousands of particles. The primary bottleneck for such experiments is the Geant4~\citep{agostinelli2003geant} simulator, which is highly CPU-demanding. Since the simulations of individual particles are independent of each other, they can be run in parallel without communication between processes. In our experiments, we split up each simulation into chunks of 2000 particles, which resulted in run times of 5-15 minutes per simulation on single CPU core, depending on the exact hardware.

\subsection{Number of simulations} \label{sup:num_sims}
Since our main idea is to train a policy for efficient optimization of downstream tasks, it is natural to ask whether the training costs are worth the efficiency gains. In our setup, we aim to minimize the number of simulator calls, since these are often the most expensive part of simulator optimization. Here, we provide an empirical upper bound on the number of `simulator calls' (as defined in section 4 of the main text) for the wireless and physics settings. Our policy is trained for a number of iterations $K$, using $G$ episodes per iteration, each of which use at most $L$ simulator calls (following the notation of Algorithm 1). Thus, we will at most employ $KGL$ simulator calls for policy training. In practice, we do not need the full $K$ iterations to train the policy (e.g., see how the sum of rewards stabilizes in Figure 12), and episodes will not use the full $L$ simulator calls (as shown by the ANC metric); $KGL$ is an upper bound. For the particle physics setting, we use $K=5$, $G=8$, and $L=30$, which equates to 1200 simulator calls to train the policy. Each simulator call involves a (parallel) computation of $5\cdot10^4$ events, so training the policy amounts to $6\cdot10^7$ events. This is more than $10\times$ less than the $10^9$ reported for optimizing a \emph{single input distribution} by \citet{massoli2015xenon1t, baranov2017optimising}. Training a policy seems to be very worthwhile in this setting. For the wireless setting, we originally use $K=20$, $G=10$, and $L=50$, which amounts to $10^4$ simulator calls. Each call involves $3000$ parallel events, so $3\cdot10^7$ events total. These numbers were chosen for expedience and are not optimized. Inspecting our training runs, we have noticed that the policy only requires (on average) 10 simulator calls for termination in the wireless setting. With $L=15$ the number of total calls drops to $3000$. The policy converges rapidly and reaches its final performance already after only 10 iterations. Setting $K=10$ further reduces the number of simulator calls to 1500. Table 1 shows that LGSO needs 23 simulator calls to optimize this setting, and our policies need 7 ($\pi$-E) or 8 ($\pi_{AL}$-E). Thus, the break-even point is at around $1500 / (23-8) = 100$ optimizations. 

\clearpage

\end{document}